\documentclass[10pt,twocolumn,letterpaper]{article}

\usepackage[pagenumbers]{cvpr}  %Required [review]
\usepackage{times}  %Required
\usepackage{helvet}  %Required
\usepackage{courier}  %Required
\usepackage{url}  %Required
\usepackage{graphicx}  %Required
\usepackage{amsmath}
\usepackage{courier}
\usepackage{amssymb}% http://ctan.org/pkg/amssymb
\usepackage{pifont}% http://ctan.org/pkg/pifont
\usepackage{soul}
\usepackage{bbm}
\usepackage{enumitem}
\usepackage{tabularx,tabulary}
\usepackage{algorithm}% http://ctan.org/pkg/algorithms
\usepackage{algpseudocode}% http://ctan.org/pkg/algorithmicx
\usepackage{amsfonts}
\usepackage{amsmath,amssymb} % define this before the line numbering.
\usepackage[table,dvipsnames,svgnames,x11names]{xcolor}
\usepackage{multirow}
\usepackage{booktabs}
\usepackage{pdfpages}
\usepackage{amsmath}
\usepackage{bbm}
\usepackage{rotating}
\usepackage[accsupp]{axessibility}

\definecolor{applegreen}{rgb}{0.55, 0.71, 0.0}
\definecolor{darkcyan}{rgb}{0.0, 0.55, 0.55}
\definecolor{darkpastelgreen}{rgb}{0.01, 0.75, 0.24}

\usepackage[pagebackref=true,breaklinks=true,colorlinks,citecolor=darkpastelgreen,bookmarks=false]{hyperref}

\usepackage{makecell}
\usepackage{amsmath}
\usepackage{bm}

\definecolor{dgreen}{rgb}{0.04,0.7,0.13}
\definecolor{maroon}{rgb}{0.75,0.07,0.03}
\definecolor{dblue}{rgb}{0.1,0.07,0.75}
\newcommand{\cmark}{{\color{dgreen}\ding{51}}}%
\newcommand{\cmarkred}{{\color{maroon}\ding{51}}}%
\newcommand{\xmark}{{\color{maroon}\ding{55}}}%
\newcommand{\xmarkgreen}{{\color{dgreen}\ding{55}}}%
\newcommand{\redcolor}[1]{\textcolor{red}{#1}}

\usepackage{epsfig}

\RequirePackage{fontawesome}
\algnewcommand\algorithmicforeach{\textbf{for each}}
\algdef{S}[FOR]{ForEach}[1]{\algorithmicforeach\ #1\ \algorithmicdo}

\usepackage{array}
\newcolumntype{P}[1]{>{\centering\arraybackslash}p{#1}}

\algdef{S}[FOR]{ForEach}[1]{\algorithmicforeach\ #1\ \algorithmicdo}

\usepackage{xcolor}% http://ctan.org/pkg/xcolor
\usepackage{algpseudocode}% http://ctan.org/pkg/algorithmicx
\usepackage{algorithm}% http://ctan.org/pkg/algorithm
\algnewcommand{\algorithmicgoto}{\textbf{go to}}%
\algnewcommand{\Goto}[1]{\algorithmicgoto~\ref{#1}}%
\definecolor{dark_green}{rgb}{0.00, 0.8, 0.0}
\definecolor{dgreen}{rgb}{0,0.35,0}

\def\cvprPaperID{10976} % *** Enter the CVPR Paper ID here
\def\confName{CVPR}
\def\confYear{2022}

\begin{document}

%%%%%%%%% TITLE

\title{Uncertainty-Aware Adaptation for Self-Supervised 3D Human Pose Estimation}

\author{Jogendra Nath Kundu$^1$ \qquad Siddharth Seth$^1$\thanks{equal contribution.} \qquad Pradyumna YM$^1$\footnotemark[1] \qquad Varun Jampani$^2$ \\Anirban Chakraborty$^1$ \qquad R. Venkatesh Babu$^1$  \\
$^1$Indian Institute of Science, Bangalore  \qquad $^2$Google Research\\
%{\tt\small jogendrak@iisc.ac.in, nav.naveenvenkat@gmail.com, rmvenkat@andrew.cmu.edu, venky@iisc.ac.in}
}

\maketitle
\thispagestyle{empty}

%%%%%%%%% ABSTRACT
\begin{abstract}
The advances in monocular 3D human pose estimation are dominated by supervised techniques that require large-scale 2D/3D pose annotations. Such methods often behave erratically in the absence of any provision to discard unfamiliar out-of-distribution data. To this end, we cast the 3D human pose learning as an unsupervised domain adaptation problem. We introduce MRP-Net\footnote{Project page: \url{https://sites.google.com/view/mrp-net}} that constitutes a common deep network backbone with two output heads subscribing to two diverse configurations; a) model-free joint localization and b) model-based parametric regression. Such a design allows us to derive suitable measures to quantify prediction uncertainty at both pose and joint level granularity. While supervising only on labeled synthetic samples, the adaptation process aims to minimize the uncertainty for the unlabeled target images while maximizing the same for an extreme out-of-distribution dataset (backgrounds). Alongside synthetic-to-real 3D pose adaptation, the joint-uncertainties allow expanding the adaptation to work on in-the-wild images even in the presence of occlusion and truncation scenarios. We present a comprehensive evaluation of the proposed approach and demonstrate state-of-the-art performance on benchmark datasets.

% The unsupervised alignment is performed in a novel way by enforcing preservation of 

\end{abstract}

%%%%%%%%% BODY TEXT

\vspace{-2mm}
\section{Introduction}
3D human pose estimation forms a core component of several human-centric technologies
% Machine perception of a 3D human pose from a monocular RGB image has gained substantial research attention. An efficient pose estimation network is often a primary requirement while developing human-centric deployable systems 
such as augmented reality \cite{hagbi2010shape}, gesture recognition~\cite{5206523}, etc. Most of the 3D human pose estimation approaches heavily rely on fully supervised training objectives \cite{Dabral_2018_ECCV, moreno20173d, sun2018integral}, demanding access to large-scale datasets with paired 3D pose annotation. However, the inconvenience of 3D pose acquisition 
% pipeline 
stands as a significant bottleneck. Unlike a 2D pose, it is difficult to manually annotate an anthropomorphically constrained 3D pose for an in-the-wild RGB image. Thus, most of the paired 3D pose datasets are collected in lab environments via body-worn sensors or multi-camera studio setups \cite{ionescu2013human3, sigal2010humaneva} that are difficult to install outdoors. This often limits the dataset diversity in terms of the variety in poses, appearances (background and lighting conditions), and outfits.

%%%%%%%%%%%%%%%%% Pose Apparel transfer %%%%%%%%%%%%%%%%%
\begin{figure}[!t]%[!tbp][h!]
\begin{center}
	\includegraphics[width=1.00\linewidth]{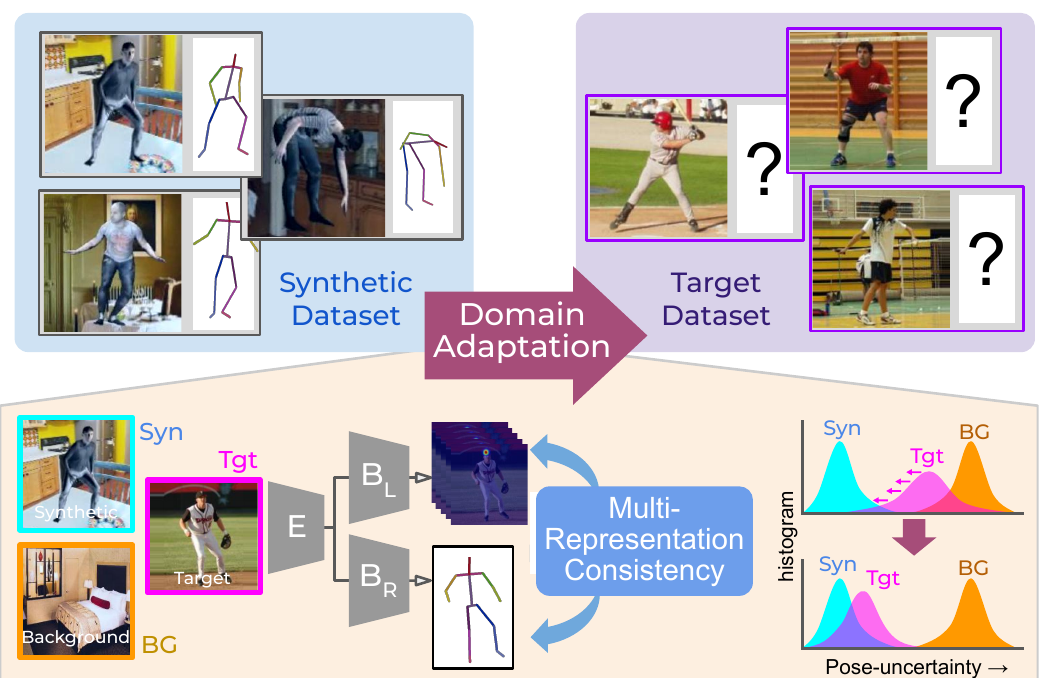}
	\vspace{-6mm}
	\caption{ 
	The proposed unsupervised adaptation framework utilizes a multi-representation consistency based uncertainty estimation for simultaneous OOD detection and adaptation.
	}
    \vspace{-7mm}
    \label{fig:concept}  
\end{center}
\end{figure}
%%%%%%%%%%%%%%%%%%%%%%%%%%%%%%%%%%%%%%%%%%%%%%%%%%%%%

Some works~\cite{rakesh2021aligning, rhodin2018learning} propose weakly supervised techniques to bypass the requirement of 3D pose annotations.
% have been proposed as a rescue. 
Several of these works leverage available paired 2D pose datasets or off-the-shelf image-to-2D 
%pose models
pose estimation networks~\cite{kong2019deep, novotny2019c3dpo, tung2017adversarial}.
%\vjnote{Add example citations}
To address the inherent 2D-to-3D ambiguity, some works either rely on multi-view image pairs~\cite{chen2019weakly, kocabas2019self, kundu2018object} or utilize unpaired 3D pose samples~\cite{chen2019unsupervised, wandt2019repnet}. % to constrain the solution space. 
Though such methods perform well when evaluated on the same dataset, %which is used for training, 
they lack cross-dataset generalization.

%\textbf{Importance of self-adapting frameworks:} 
Consider a scenario where we want to deploy a 3D human pose estimation system in a new application environment
% deployment of a pose estimation system in a new application environment 
(\ie target domain). From the vendor’s perspective, a general approach would be to improve the system’s generalization via supervised training on a wide variety of labeled source domains~\cite{li2017deeper}. However, target-specific 
%\vjnote{Do you mean target-specific? 'environment' is ambiguous.} 
training usually achieves the best performance beyond the generic system. Though it is not convenient to collect annotations for every novel deployment scenario, an effective unsupervised adaptation framework stands as the most practical way forward. Unsupervised adaptation~\cite{ganin2015unsupervised,tzeng2017adversarial, kundu2019_um_adapt} seeks a learning technique that can minimize the domain discrepancy between a labeled source and an unlabeled target. Thus, the vendor has to collect unlabeled RGB inputs from the new environment to enable the adaptation process. Let us consider a different scenario where the target environment is identical to one of the source domains implying no domain-shift. Here, the vendor can choose to directly deploy the generic system without adaptation training. However, the system must have a provision to detect whether it is required to run the adaptation process. In other words, it should have the ability to discern out-of-distribution (OOD) scenarios~\cite{hendrycks2016baseline,liang2017enhancing,lee2018simple}. Such an ability is more crucial while deploying in a continually changing environment~\cite{wulfmeier2018incremental}, \eg a model adapted for sunny weather conditions 
% based on the deployment-time weather conditions 
would fail while encountering rainy weather, thus requiring re-adaptation.

%%%%%%%%%%%%%%%%%%%% Characteristic comparison starts %%%%%%%%%%%%%%%%%%%%%%%%%%%%%
\begin{table}[t]
	\footnotesize
	\caption{
	Comparison of positive (in green) and negative (in red) attributes of ours against prior 3D human pose estimation methods. %Sup. stands for supervision. %Comparison of our approach against prior human 3D pose estimation works, in terms of access to supervision and generalizability in various scenarios. 
	}
	\vspace{-2mm}
	\centering
	\setlength\tabcolsep{2.0pt}
	\resizebox{0.47\textwidth}{!}{
	\begin{tabular}{l|ccc|c|cc}
	\hline
 		\multirow{3}{*}{Methods} & \multicolumn{3}{c|}{ \makecell{Real Sup.}} &
 		%\multicolumn{2}{c|}{\makecell{Unpaired sup}} &
 		\multirow{3}{*}{\makecell{Synthetic\\3D pose\\ Sup.}} &
 		\multicolumn{2}{c}{\multirow{2}{*}{ \makecell{Generalization\\capability}}} \\
 		\cline{2-4}%\\
 		 & \multirow{2}{*}{\makecell{Multi\\view}} & \multirow{2}{*}{\makecell{2D\\pose}} & \multirow{2}{*}{\makecell{3D\\pose}}  & & \\
 		 \cline{6-7} &&&&& \makecell{Occlusion} & \makecell{Uncertainty}%\makecell{2D \\pose} & \makecell{3D \\pose} 
 		 \\ \hline\hline
  		\rowcolor{gray!00}
  		Zhou \etal~\cite{7780906} &\xmarkgreen &\cmarkred &\xmarkgreen &\xmarkgreen &\xmark & \cmark %($\sim$4k)
 		\\
  		
 		Rhodin \etal~\cite{rhodin2018unsupervised} &\cmarkred &\xmarkgreen &\xmarkgreen &\xmarkgreen &\xmark &\xmark %($\sim$4k)
 		\\
 		
 		Iqbal \etal~\cite{Iqbal_2020_CVPR} &\cmarkred &\cmarkred &\xmarkgreen &\xmarkgreen &\xmark &\xmark\\
 		Doersch \etal~\cite{NIPS2019_9454} &\xmarkgreen &\xmarkgreen &\xmarkgreen &\cmark &\xmark &\xmark    \\
 		LCR-Net++~\cite{rogez2019lcr} &\xmarkgreen &\cmarkred &\cmarkred &\xmarkgreen &\cmark &\xmark    \\%\hline
 		PoseNet3D~\cite{tripathi2020posenet3d} &\xmarkgreen &\cmarkred &\xmarkgreen & \xmarkgreen & \xmark & \xmark \\
  		% \rowcolor{gray!14}
 		\textit{Ours} &\xmarkgreen &\xmarkgreen &\xmarkgreen &\cmark &\cmark &\cmark   \\ 
		\hline
	\end{tabular}}
	\vspace{-4mm}
	\label{tab:char}
\end{table} 
%%%%%%%%%%%%%%%%%% Characteristic comparison ends  %%%%%%%%%%%%%%%%%%%%%%%%%%

We propose a novel domain adaptation (DA) framework, \textit{MRP-Net} (Fig.~\ref{fig:concept}), equipped with uncertainty estimation~\cite{kuleshov2018accurate} for the monocular 3D human pose estimation task. 
%Though this dataset encapsulates a wide range of diversity, the resultant model still suffers from poor generalization on natural images due to the synthetic-real domain-shift. 
To this end, we use a multi-representation pose network with a common backbone followed by two pose estimation heads subscribing to two diverse % design 
output configurations; a) Heat-map based joint localization and b) Model-based parametric regression. This not only encourages ensemble-diversity required for uncertainty estimation~\cite{gal2016dropout} but also allows us to encompass the merits of both schools of thought~\cite{sun2018integral,moon2020i2l}. The former configuration advocates maintaining the spatial structure via a fully-convolutional design~\cite{Mehta2017vnect,newell2016stacked,sun2019deep} while lacking provisions to inculcate structural articulation and bone-length priors. The latter advocates regressing a parametric form of the pose as a whole (via fully-connected layers)~\cite{Li_2020_CVPR,TrumbleBMVC2017,kanazawa2018end,kolotouros2019learning} while allowing model-based structural prior infusion~\cite{pavllo2018quaternet, Kundu_2020_WACV}.
We use the 3D graphics-based synthetic SURREAL dataset \cite{varol2017learning} as the labeled source domain to supervise our backbone network.

In addition, we derive useful measures to quantify the prediction uncertainty at two granularity levels; viz a) \textit{pose-uncertainty}, b) \textit{joint-uncertainty}. %Pose-uncertainty is defined as the disagreement between predictions through the two ensemble heads. Whereas joint-uncertainty is expressed as the entropy of the joint-wise heat-map distributions. 
During training, we utilize both a labeled source and a dataset of backgrounds (BG) to elicit the desired behavior of the 
% defined 
uncertainties. %by maximizing the uncertainty for BG while minimizing it for the source. 
Here, the backgrounds approximate an extreme out-of-distribution scenario. Upon encountering the unlabeled target, the adaptation process seeks to reduce the target uncertainties alongside a progressive self-training on a set of reliable pseudo-labels. Alongside the adaptation for datasets with full-body visibility, the \textit{joint-uncertainty} lays a suitable ground to expand our adaptation to work on in-the-wild target domain (unlabeled) with partial body visibility (\ie under external occlusion or truncated frame scenarios). We present an extensive evaluation of the proposed framework under a variety of source-to-target %domain adaptation 
settings. %Fig.~\ref{fig:concept} gives an overview of our approach.
%We perform extensive experiments to validate the efficacy of our approach and demonstrate superior generalizability on samples from unseen wild environments. 
In summary: %our contributions are as follows:
\begin{itemize}
\vspace{-1.8mm}
\item %We propose a novel domain adaptation framework, MRP-Net, equipped with uncertainty estimation that uses a multi-representation pose network. Here, \textit{pose-uncertainty} is quantified as the disagreement in pose predictions obtained through the two output heads subscribing towards two diverse design configurations (model-free versus model-based).
We propose a novel domain adaptation framework, \textit{MRP-Net}, that uses a multi-representation pose network. % equipped with uncertainty estimation. 
Here, \textit{pose-uncertainty} is quantified as the disagreement between pose predictions through the two output heads subscribing towards two diverse design configurations (model-free versus model-based).

\vspace{-1.8mm}
\item We propose to utilize negative samples (backgrounds and simulated synthetic joint-level occlusions) to improve the effectiveness of the proposed pose and joint uncertainties. The presence of negatives also helps to retain the uncertainty estimation ability even while adapting to a novel target scenario.

\vspace{-1.8mm}
\item %We demonstrate state-of-the-art 3D human pose estimation performance among the semi/self-supervised prior arts on Human3.6M~\cite{ionescu2013human3}. 
Our synthetic (SURREAL) to in-studio adaptation outperforms the comparable prior-arts on Human3.6M~\cite{ionescu2013human3}. Our in-studio (Human3.6M) to in-the-wild adaptation achieves state-of-the-art performance across four datasets. 
% 3DHP~\cite{mehta2017monocular}, and 3DPW~\cite{von2018recovering}. 
% %\vjnote{Add citations}
We show uncertainty-aware 3D pose estimation results for unsupervised adaptation to in-the-wild samples with partial body visibility.

%which establishes the generalizability of our approach to in-the-wild samples with partial body visibility.

% We show impressive \vjnote{`impressive' is subjective and that is up to the reader to decide. Just say that our approach generalizes well to in-the-wild samples with partial body visibility?} uncertainty-aware 3D pose estimation results for unsupervised adaptation to in-the-wild samples with partial body visibility.
\end{itemize}

% %%%%%%%%%%%%%%%%%%%%%%%%%%%%%%%%%%%%%%%%%%%%%%%%%%%%%       _                    _        
% |  __ \    | |     | |         | |                  | |       
% | |__) |___| | __ _| |_ ___  __| |_      _____  _ __| | _____ 
% |  _  // _ \ |/ _` | __/ _ \/ _` \ \ /\ / / _ \| '__| |/ / __|
% | | \ \  __/ | (_| | ||  __/ (_| |\ V  V / (_) | |  |   <\__ \
% |_|  \_\___|_|\__,_|\__\___|\__,_| \_/\_/ \___/|_|  |_|\_\___/
% %%%%%%%%%%%%%%%%%%%%%%%%%%%%%%%%%%%%%%%%%%%%%%%%%%             

\section{Related Works}
\label{sec:related-works}
%\input{related_works}
%\ssnote{Reference Table-1 in the related works.}

Table~\ref{tab:char} shows a comparison of our approach against related prior approaches. Here, Sup. stands for supervision.
%Table~\ref{tab:char} shows a comparison of our approach against the a set of related prior 3D human pose estimation approaches.

%\textbf{Model free and model based.}
\noindent
\textbf{Domain Adaptation.} Cao~\etal~\cite{cao2019cross} propose to apply discriminator based discrepancy minimization technique for the animal pose estimation task. %There are a number of works \cite{Deep3DPose,7301337,NIPS2019_9454} that use synthetic data for training 3D human pose estimation networks. 
To address the synthetic-to-real domain gap for 3D human pose estimation, Doersch~\etal \cite{NIPS2019_9454} propose to use optical-flow and 2D key-points as the input as these representations are least affected by domain shift unlike RGB images (texture and lighting variations). Similarly, Zhang~\etal~\cite{mm_domain19} propose to leverage multi-modal input, such as depth and body segmentation masks. Mu~\etal~\cite{9157335} leverage several consistency losses to effectively adapt from source to target.
%to adapt for the 3D human pose estimation task. 
Our proposed framework does not access any such auxiliary input modality. Recently, some works~\cite{syguan2021boa,inference_stage2020} propose online test-time adaptation of 3D human pose estimation from in-studio source to in-the-wild target.

%in order to adapt the 3D human pose systems to changing scenarios.

\noindent
\textbf{Pose estimation in presence of occlusion.} In literature, we find some methods that address human pose estimation in presence of partial occlusion. Several works design techniques to estimate location of the occluded keypoint conditioned on the unoccluded ones while accessing additional spatio-temporal~\cite{radwan2013monocular,rogez2017lcr,de2018deep,cheng2019occlusion,cheng20203d} or scene related context~\cite{zanfir2018monocular,zanfir2018deep, kundu2020unsup}. Mehta~\etal~\cite{mehta2018single} propose to use occlusion-robust pose-maps to address partial occlusion scenarios. % within the image frames.

\noindent
\textbf{Monocular 3D human pose estimation.}
In literature, we find two broad categories 
%Single-image 3D pose estimation methods are studied under two broad categories 
viz. a) methods that directly infer the 3D pose representation \cite{Rosales2001LearningBP, 1542030, 5206699} and b) methods using model-based parametric representation \cite{6909612, 6619308, SMPL-X:2019, Bogo:ECCV:2016, kundu_appearance}. The former directly maps the input image to the 3D pose while the latter maps images to latent parameters of a predefined parametric human model. The latter setup provides a suitable ground to impose the kinematic pose priors via adversarial training~\cite{kundu2021non, kundu2020cross}. %However, the former yields superior performance it enjoys strong gradient back-propagation in the absence of a predefined parametric model. 
The former setup is further %Such approaches can be further 
categorized into one-stage \cite{Zhou_2017_ICCV, sun2018integral, 8099622, 8237635, pavlakos2018ordinal, 8578649} and two-stage methods \cite{zhaoCVPR19semantic, martinez2017simple, moreno20173d, 10.1007/978-3-030-01249-6_5}. One-stages approaches directly map images to the 3D poses. Whereas, two-stage methods first map images to an 2D pose representation followed by another mapping to perform the 2D-to-3D lifting. 
%In our work, the shared latent pose can be seen as a parametric form to represent plausible 3D poses. And, the \textit{image-to-latent} model is trained to regress the latent pose parameters with latent being an intermediate representation which is not related to 2D poses.

\noindent
\textbf{Pose estimation via multi-head architecture.\!} {\fussy PoseNet3D\ \cite{tripathi2020posenet3d} employs a student-teacher multi-head framework. However, the primary task is 2D-to-3D lifting where they rely on 2D pose predictions obtained from fully supervised image-to-2D pose model~\cite{newell2016stacked}. Unlike PoseNet3D, we do not leverage in-the-wild 2D pose annotations or temporal consistency. Further, prior arts~\cite{Guler2018DensePose, Guler_2019_CVPR} also employ similar multi-head architecture to leverage auxiliary supervision or to improve predictions through consistency losses. To the best of our knowledge, none of the prior-arts utilize such architecture for OOD or self-adaptation to unlabeled target.}

%%%%%%%%%%%%%%%%% Pose EnGAN Fig. %%%%%%%%%%%%%%%%%
\begin{figure*}%[!tbp]%[!b]%[!tbp][h!]
\begin{center}
    %\vspace{-4mm}
	\includegraphics[width=1.0\linewidth]{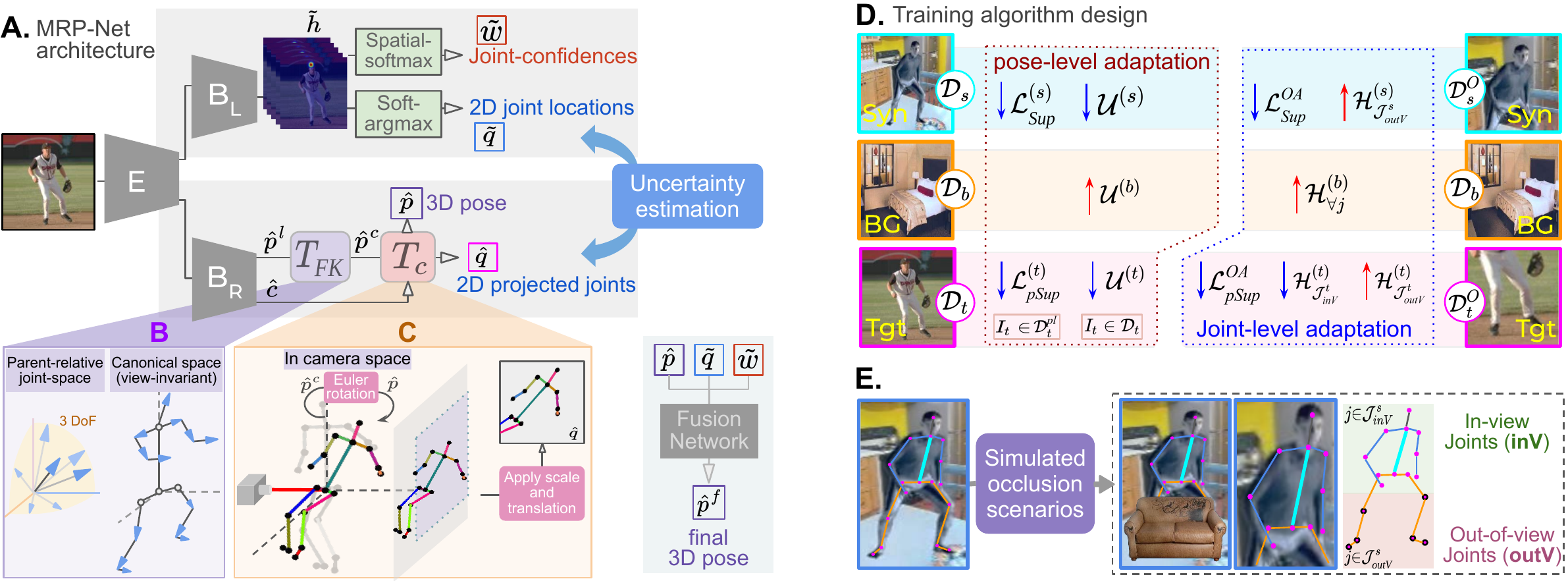}\vspace{-2mm}
	\caption{
	An overview of the proposed framework. \textbf{A.} Design configuration for output representations of \textit{MRP-Net} architecture.
	\textbf{B.} Details of the Forward-kinematics transformation.
	\textbf{C.} Applying rotation and camera transformations.
	\textbf{D.} An illustration of the datasets and loss terms for the proposed pose-level and joint-level adaptation.
	\textbf{E.} The occlusion simulation to obtain in-view and out-view joint-ids.
	}
 	\label{fig:main1}    
    \vspace{-5mm}
\end{center}
\end{figure*}

%**********************************************
%%%%%%%%%%%%%%%%%%%%%%%%%%%%%%%%%%%%%%%%%%%%%%%%%%%%%

%% https://www.askapache.com/online-tools/figlet-ascii/ (big) %%%
%     /\                                    | |    
%    /  \   _ __  _ __  _ __ ___   __ _  ___| |__  
%   / /\ \ | '_ \| '_ \| '__/ _ \ / _` |/ __| '_ \ 
%  / ____ \| |_) | |_) | | | (_) | (_| | (__| | | |
% /_/    \_\ .__/| .__/|_|  \___/ \__,_|\___|_| |_|
%          | |   | |                               
%          |_|   |_|                               
%%%%%%%%%%%%%%%%%%%%%%%%%%%%%%%%%%%%%%%%%%%%%%%%%%

\vspace{-1mm}
\section{Approach}
\label{sec:approach}
We aim to prepare a pose estimation network that can discern OOD samples by delivering a high prediction uncertainty for such inputs. Simultaneously, the network should not compromise on pose estimation performance for in-domain inputs. Sec. \ref{subsec:architecture} first discusses the pros and cons of the two widely used design configurations specific to output representation of human pose estimation networks. We describe the key design components of the proposed \textit{MRP-Net} architecture, following which we propose intuitive ways to quantify the pose and joint uncertainties. Sec. \ref{subsec:poseleveladaptationframework} illustrates the training procedure to progressively strengthen and leverage the \textit{pose-uncertainty} for the unsupervised DA setting. In Sec \ref{jointleveladaptationframework}, we leverage the \textit{joint-uncertainties} as a means to expand the adaptation to a broader scope, \ie to in-the-wild targets in the presence of occlusion and truncations.

\subsection{Pose estimation architecture} 
\label{subsec:architecture}
%Our goal is to design a suitable architecture for 3D human pose estimation with instilled uncertainty estimation capability.  Pose estimation architectures employ a wide variety of design configurations. %to regress or localize the 3D/2D pose representations. One can group these into the following two broad categories.

In literature, pose estimation architectures employ one of the following two  design configurations.

\vspace{0.7mm}
\noindent \textbf{a) Localization-based representation.} 
Most of the popular 2D pose estimation approaches employ fully convolutional architectures (such as hourglass networks~\cite{newell2016stacked}), where the final pose is realized via $J$ heatmaps, one for each joint~\cite{Mehta2017vnect}. Here, the heatmaps are treated as spatial probability distributions (PDFs) with a probability peak near the spatial joint location. This can also be viewed as a localization based model-free design as it refrains from utilizing the joint-connectivity and the bone-length knowledge.

\vspace{0.7mm}
\noindent \textbf{b) Regression-based representation.} Here, networks aim to directly regress joint coordinates or some rich parametric representations (latent) of the final pose~\cite{kanazawa2018end,TrumbleBMVC2017}. Networks employ fully-connected layers after the back-bone CNN, thereby breaking away from the spatial structure to learn a highly non-linear mapping, unlike the localization-based design. One can easily inculcate joint-connectivity or bone-length priors via model-based design with integrated forward kinematics~\cite{kundu2020ksp,zhou2016deep,pavllo2018quaternet}. However, such model-based representation does not allow a provision to extract joint-level uncertainty as it sees the pose as a whole.

%\vspace{0.7mm}
%\noindent \textbf{Requirements for uncertainty estimation.}
%As pose estimation models are being deployed for safety-critical tasks, such as human motion prediction in self-driving cars, medical alert systems for abnormality in patient’s posture, etc., uncertainty estimation becomes more crucial to gauge the severity of emergency situations. 
%Any deployable system may get exposed to unfamiliar inputs either due to a domain-shift or anomalous out-of-distribution (OOD) data. 
Normally trained systems often behave erratically in the absence of any provision to discard out-of-distribution inputs. In literature, ensemble-based systems~\cite{lakshminarayanan2016simple} %cite% https://openreview.net/pdf?id=BygSP6Vtvr
have been used to derive useful uncertainty measures. Several approaches resort to random initialization or dataset bootstrapping to induce \textit{ensemble-diversity} which is crucial to realize a robust uncertainty quantification metric.

\vspace{-3mm}
\subsubsection{{MRP-Net} architecture} 
\vspace{-1mm}
Keeping in mind the computational overhead of full network ensembles, we decide to develop multi-head ensembles with a common CNN backbone. As shown in Fig.~\ref{fig:main1}, the multi-head ensemble consists of a common encoder backbone ${E}$ which is followed by two ensemble heads that are denoted as ${B}_L$ and ${B}_R$. Unlike the random initialization strategy, we propose to maintain ensemble diversity by following the above discussed pose modeling configurations.

\vspace{0.8mm}
\noindent
\textbf{a) Joint-localization at ${B}_L$ output.}
The localization branch ${B}_L$ is a convolutional decoder which outputs heatmap PDFs, $\tilde{h}:\{\tilde{h}^{(j)}\}_{j=1}^J$ (via spatial-softmax). %with $J$ being the total number of joints. 
These heatmaps are converted to 2D joint coordinates %$\mathbb{R}^{J\times 2}$ 
via a soft-argmax operation, $\tilde{q}^{(j)} = \sum_{v}(v)h^{(j)}(v)$. Here, $v:[v_x,v_y]$ denotes the spatial grid index. We also extract joint-confidence, $\tilde{w}$ as $\tilde{w}^{(j)}=\max_{v}h^{(j)}(v)$. %On the other hand, the regression branch ${B}_R$ consists of several fully-connected layers to regress a 3D pose parameterization, $\hat{p}^l$ and camera-parameters, $\hat{c}$.

\vspace{0.8mm}
\noindent
\textbf{b) Kinematic-parameterization at ${B}_R$ output.}
On the other hand, the regression branch ${B}_R$ consists of several fully-connected layers to regress a 3D pose parameterization, $\hat{p}^l$ and camera-parameters, $\hat{c}$. We design a simple kinematic model based on the knowledge of hierarchical limb connectivity and relative bone-length ratios. We aim to disentangle the rigid camera variations (in camera space) from the non-rigid limb articulations (in canonical space). Here, the non-rigid articulations are modeled at the view-independent canonical space. Note that, in canonical space, the pelvis joint exactly aligns with the origin while the skeleton faces towards the %+ve 
{positive} $X$-axis, thus making it a view-independent pose representation, $\hat{p}^c\in \mathbb{R}^{J\times 3}$. In our convention, the skeleton-face is obtained as the cross-product direction of two vectors; i.e., left-hip to neck and left-hip to right-hip. However, directly regressing the canonical pose coordinates $\hat{p}^c$ does not ensure the 3D bone-length constraints. Thus, we obtain the canonical pose via a forward-kinematic transformation where the pose-network regresses local limb vectors of unit magnitudes, $\hat{p}^l\in\mathbb{R}^{J\times 3}$. For each joint $j$, the limb-vector is defined at a predefined convention of parent-relative joint space. Here, the forward-kinematic transformation ${T}_\textit{FK}$ builds the canonical pose by recursively traversing over joints in the kinematic tree; while applying pre-fixed bone-length magnitudes along the transformed limb-vector directions (see Fig. \ref{fig:main1}\redcolor{B}). Alongside the local limb-vectors, the pose-network regresses the Euler-rotations alongside the scale and spatial translation parameters ($\hat{c}$: 3 angles, 1 scale, and 2 translation parameters). Following this, scaled orthographic projection ${T}_c$ outputs the projected image-space joint coordinates $\hat{q}\in\mathbb{R}^{J\times 2}$. This also outputs the camera-space 3D pose $\hat{p}\in\mathbb{R}^{J\times 3}$ as an intermediate representation.

\vspace{1mm}
\noindent
Next, we quantify uncertainty as follows:\\
\noindent \textbf{a) Quantifying pose-level uncertainty.}
In literature, ensemble disagreement provides a useful quantitative measure to evaluate the prediction uncertainty~\cite{gal2016dropout}. % For discrete output space KL-divergence between the posteriors of the two ensemble networks is treated as a measure of uncertainty.
For \textit{MRP-Net}, we propose to rely on the diversity in design configuration between the two representations obtained via ${B}_L$ and ${B}_R$. %Thus, the prediction uncertainty for pose estimation is quantified via a consistency measure between the outputs, obtained via ${B}_L$ and ${B}_R$. 
Thus, we define the \textit{pose-uncertainty} as follows:

\vspace{-7mm}
\begin{equation}
    \begin{aligned}
    \small
    \boldsymbol{\mathcal{U}(I) = |\tilde{q}-\hat{q}|};\;\; \tilde{q}=B_L\!\circ\! {E}(I),\;\; \hat{q}={T}\!\circ\! B_R\circ {E}(I) 
    \end{aligned}
\end{equation}

\vspace{-2mm}
% \sum_{j=1}^J(1-\tilde{w}^{(j)})
\noindent Here, $\circ$ denotes functional composition and ${T}={T}_\textit{FK}\circ {T}_c$.  %$\tilde{q}=B_L\circ {E}(I)$, and $\hat{q}={T}_\textit{FK}\circ {T}_c\circ B_R\circ {E}(I)$. We weight the 

\vspace{1mm}
\noindent  \textbf{b) Quantifying joint-level uncertainty.}
Among the two representations, joint-level uncertainty can be extracted from localization based spatial map distributions. For each joint prediction, the joint uncertainty associated with a joint \textit{j}, is realized as the self-entropy of spatial distributions, \ie 
\vspace{-2.5mm}
\begin{equation}
\begin{aligned}
\boldsymbol{\mathcal{H}(I, j)} = -\sum_{v}\tilde{h}^{(j)}(v)\log \tilde{h}^{(j)}(v)
\end{aligned}
\end{equation}

\vspace{-6mm}
\subsection{Pose-level adaptation framework}
\label{subsec:poseleveladaptationframework}

In unsupervised DA the primary goal is to transfer the task knowledge from a labeled source dataset $\boldsymbol{\mathcal{D}_s}$ (synthetic domain) to an unlabeled target dataset $\boldsymbol{\mathcal{D}_t}$ (real domain). 

%For the monocular pose estimation task, we use \textit{SURREAL} as the synthetic source dataset $\boldsymbol{\mathcal{D}_s}$. Let $\boldsymbol{\mathcal{D}_t}$ denote the unlabeled target dataset consisting of images from a real deployed environment. %The synthetic supervision loss is denoted as;
%$$
%\mathcal{L}_\textit{Sup}^{(s)}(I\in\mathcal{D}_s)={L}_{h}(\tilde{h}, h_\textit{gt}) + \lambda |\hat{p}-p_\textit{gt}|
%$$

% \noindent
% \textbf{Preparing uncertainty-aware \textit{MRP-Net}.}

\vspace{-4mm}
\subsubsection{Preparing pose-uncertainty-aware \textit{MRP-Net}} \vspace{-1mm}
Let, $\mathcal{L}_h(.)$ and $\mathcal{L}_p(.)$ be the mean squared loss for the heatmap and the 3D pose respectively.  %We first train the network to minimize the following 
The synthetic supervision loss is expressed as;

% \begin{equation}\label{eq2}
%     \boldsymbol{\mathcal{L}_\textit{Sup}^{(s)}}(I\in\mathcal{D}_s)=\mathcal{L}_{h}(\tilde{h}, h_\textit{gt}) + \lambda_1 \mathcal{L}_p(\hat{p},p_\textit{gt}) + \lambda_2\mathcal{U}^{(s)}
% \end{equation}
\vspace{-6mm}
\begin{equation}
\begin{aligned}
\small
\boldsymbol{\mathcal{L}_\textit{Sup}^{(s)}}(I\in\mathcal{D}_s)=\mathcal{L}_{h}(\tilde{h}, h_\textit{gt}) + \lambda_1 \mathcal{L}_p(\hat{p},p_\textit{gt}) + \lambda_2\mathcal{U}^{(s)}
\end{aligned}
\end{equation}

\noindent Here, $h_\textit{gt}$ and $p_\textit{gt}$ denote the respective ground-truths (GT) with $\lambda_1$ and $\lambda_2$ being the balancing hyperparameters. The intended behaviour of \textit{pose-uncertainty} is that it would elicit a high
value for $\mathcal{U}^{(s)}$ for unfamiliar inputs while being low for familiar in-domain samples, \ie for $I\in\mathcal{D}_s$. However, training of \textit{MRP-Net} solely on samples from $\boldsymbol{\mathcal{D}_s}$ outputs consistently low \textit{pose-uncertainty} for both {in-domain} and {out-of-domain} samples during validation. %To elicit the intended behaviour, 
One has to explicitly update the network parameters to obtain higher uncertainty for the unfamiliar inputs. In view of the human pose estimation task, we resort to a dataset of background images $\boldsymbol{\mathcal{D}_b}$ (\ie images without any person in frame) to approximate an extreme out-of-distribution scenario. %Thus, we train \textit{MRP-Net} to maximize $\mathcal{U}(I\in\mathcal{D}_b)$. 

%\noindent
In summary, the \textit{MRP-Net} is trained to minimize $\mathcal{L}_\textit{Sup}^{(s)}$ while simultaneously maximizing the \textit{pose-uncertainty} for backgrounds \ie $\boldsymbol{\mathcal{U}^{(b)}}= \mathcal{U}(I\in\mathcal{D}_b)$.

% \noindent
% \textbf{Adaptation via uncertainty minimization.}
\vspace{-4mm}
\subsubsection{Adaptation via uncertainty minimization}
\vspace{-1mm}
Next, the \textit{uncertainty-aware} network is exposed to the unlabeled target samples, $I_t\in\mathcal{D}_t$. Analyzing the histogram from Fig. {\color{red}1} of the \textit{pose-uncertainties} for samples from $\mathcal{D}_s$, $\mathcal{D}_b$, and $\mathcal{D}_t$ shows that the uncertainties for $\mathcal{D}_t$ spans a wide-range of values with the same for $\mathcal{D}_s$ and $\mathcal{D}_b$ as peaky distributions at opposite extremes. One can relate the uncertainty gap between the samples from $\mathcal{D}_b$ and $\mathcal{D}_t$ as result of the distinction between \textit{data-uncertainty} (or aleatoric uncertainty) and \textit{knowledge-uncertainty} (or epistemic uncertainty), respectively. Here, \textit{data-uncertainty} refers to the irreducible uncertainty in prediction as a result of noisy input, whereas the \textit{knowledge-uncertainty} refers to the reducible uncertainty elicited as an outcome of the discrepancy in input distributions (\ie the synthetic versus real domains). 

\vspace{0.5mm}
\noindent \textbf{a) Adaptation.} Motivated by the above discussion, we seek to minimize the \textit{pose-uncertainty} for the target samples, \ie $\boldsymbol{\mathcal{U}^{(t)}}= \mathcal{U}(I\in\mathcal{D}_t)$ alongside minimizing $\boldsymbol{\mathcal{L}_\textit{Sup}^{(s)}}$, while simultaneously maximizing $\boldsymbol{\mathcal{U}^{(b)}}= \mathcal{U}(I\in\mathcal{D}_b)$. 

\vspace{0.5mm}
\noindent \textbf{b) Self-training on target pseudo-labels.}
The literature~\cite{lee2013pseudo,zou2018unsupervised} suggests that target supervision on a reliable pseudo-label subset helps to improve the adaptation performance. In classification tasks, the class predictions of the most confident targets are collected as a reliable pseudo-label subset~\cite{saito2017asymmetric}. In the proposed scenario, the reliability of pseudo-label selection based on the \textit{pose-uncertainty} is highly questionable, as it might be an outcome of the enforced uncertainty minimization loss instead of a genuine learning induced behaviour. Thus, in order to move away from such dependency, we utilize an equivariance consistency based pseudo-label selection criteria. This is realized by applying the most diverse spatial transformation \ie \textit{image-flip}. Essentially, the 2D pose predictions of a given image, $I_t$ and the corresponding flipped image, $I_t^\prime=\mathcal{F}_I(I_t)$ are compared after a left-right joint-id swapping operation $\mathcal{F}_q$. Thus, for each $I_t$ we obtain the following predictions $\hat{q}_t, \tilde{q}_t, \mathcal{F}_q(\hat{q}_t^\prime), \mathcal{F}_q(\tilde{q}_t^\prime)$. Finally, the pseudo-label subset $\mathcal{D}_t^\textit{pl}$ is realized by selecting samples that have an equivariance-consistency less than a preset threshold $\alpha_p^\textit{th}$ \ie,
\vspace{-2mm}
\begin{equation}
\begin{aligned}
\mathcal{D}_t^\textit{pl} = \{I_t: (|\hat{q}_t - \mathcal{F}_q(\tilde{q}_t^\prime)| + |\tilde{q}_t - \mathcal{F}_q(\hat{q}_t^\prime)|)< \alpha_p^\textit{th}\}
\end{aligned}
\end{equation}
\noindent Next, we minimize the target pseudo-label supervision loss;
\\
\vspace{-3mm}
\begin{equation}
\begin{aligned}
\small
\boldsymbol{\mathcal{L}^{(t)}_\textit{pSup}}(I\in\mathcal{D}_t^\textit{pl}) = \sum_{j=1}^J \tilde{w}^{(j)} (\mathcal{L}_{h}^{(j)}(\tilde{h}, h^\textit{pl}_\textit{gt}) + \lambda \mathcal{L}^{(j)}_p(\hat{p},p^\textit{pl}_\textit{gt}))
\end{aligned}
\end{equation} \vspace{-2mm}

Here, the supervised joint-wise loss is weighted by the normalized joint-confidences to avoid strong supervision on confusing joint predictions. $p^\textit{pl}_\textit{gt}$ and $h^\textit{pl}_\textit{gt}$ denote the estimated pseudo-label GT (\ie prediction average over the equivariance instances). Here, $\mathcal{D}_t^\textit{pl}$ alongside the pseudo-label GTs are updated at regular intervals during training.
%periodically after a fixed training iteration intervals.

\subsection{Joint-level adaptation framework} \label{jointleveladaptationframework}
Most of the prior 3D pose estimation approaches expect full-body visibility without external occlusion. However, in real-world deployment, the camera feed may capture human images having external object occlusions or truncations. In such scenarios, an intended behavior of the model would be to estimate a reasonably well joint localization specifically for the \textit{in-view} joints with lower \textit{joint-uncertainty} values, and higher joint-uncertainties for the \textit{out-view} joints. %Here, the joint-level uncertainty (also termed as join-confidences) is realized as the self-entropy of spatial-pdfs obtained for the joint-level heat-maps at the output representation of the CNN branch.

% \vspace{1mm}
% \noindent
% \textbf{a) Preparing joint-uncertainty-aware \textit{MRP-Net}.}
\vspace{-4mm}
\subsubsection{Preparing joint-uncertainty-aware \textit{MRP-Net}} \vspace{-1mm}
Similar to \textit{pose-uncertainty}, one must simulate joint-level uncertainties to enable the model to elicit the above discussed behavior. Note that, training on synthetic full-body images (\ie $\mathcal{D}_s$) or the backgrounds (\ie $\mathcal{D}_b$) is not suitable enough as they do not encourage varying \textit{joint-uncertainties} for the same input instance. Thus, we simulate an \textit{occlusion-aware synthetic dataset}, $\boldsymbol{\mathcal{D}_s^\textit{O}}$ with segregated set of \textit{in-view} and \textit{out-view} image-joint pairs, denoted as $\mathcal{J}^s_\textit{inV}$ and $\mathcal{J}^s_\textit{outV}$ respectively (see Fig. \ref{fig:main1}\redcolor{D}). Broadly, we simulate occlusion of two kinds, viz, a) occlusion by an external object, and b) truncation of the image frame. %To avoid unreasonable or near full-body occlusions, we restrict the occlusion simulations to occlude either the upper or the lower body regions keeping at least one full limb on clear visibility. 
We apply the following synthetic supervision loss. %($\mathbbm{1}$ denotes an indicator function).
% \begin{equation}
%     \begin{aligned}
%     \small
%     \boldsymbol{\mathcal{L}_\textit{Sup}^\textit{OA}} (I\in \mathcal{D}_s^O) = \mathbbm{1}_{\mathcal{J}_\textit{inV}^s} (\mathcal{L}_h^{(j)}(\tilde{h}, h_\textit{gt}) + \lambda_1 \mathcal{L}_p^{(j)}(\hat{p}, p_\textit{gt})) \\ \small - \lambda_2 \boldsymbol{\mathcal{H}^{(s)}_{\mathcal{J}^{s}_\textit{outV}}}
%     \end{aligned}
% \end{equation}
\vspace{-2mm}
\begin{equation}
    \begin{aligned}
    \vspace{-4mm}
    \small
    \boldsymbol{\mathcal{L}_\textit{Sup}^\textit{OA}} (I\!\in\! \mathcal{D}_s^O) = \mathbbm{1}_{\mathcal{J}_\textit{inV}^s} (\mathcal{L}_h^{(j)}(\tilde{h}, h_\textit{gt}) + \lambda_1 \mathcal{L}_p^{(j)}(\hat{p}, p_\textit{gt})) - \lambda_2 \boldsymbol{\mathcal{H}^{(s)}_{\mathcal{J}^{s}_\textit{outV}}}
    \end{aligned}
\end{equation}

%((I,j)\in\mathcal{J}_\textit{outV}^s )
%$$

\noindent
Here, $\mathbbm{1}$ denotes an indicator function. The last-term aims to maximize the \textit{joint-uncertainties} only for the \textit{out-view} joints. We also maximize \textit{joint-uncertainties} of all the joints for the backgrounds $\mathcal{D}_b$ by maximizing $\boldsymbol{\mathcal{H}_{\forall j}^{(b)}}$.

% \vspace{1mm}
% \noindent
% \textbf{b) Adapting to unlabeled target with occlusion.}
\vspace{-4mm}
\subsubsection{Adapting to unlabeled target with occlusion}
\vspace{-1mm}
Next, the \textit{joint-uncertainty-aware} model is exposed to samples from the unlabeled target dataset containing images of varied kinds including full-body, truncated, and occluded samples. We denote this dataset as $\boldsymbol{\mathcal{D}_t^\textit{O}}$ against the \textit{full-body} target $\mathcal{D}_t$. We follow the adaptation process very similar to that of the proposed pose-level adaptation with the \textit{pose-uncertainties} replaced by the \textit{joint-uncertainties}. 

We follow the similar equivariance-based pseudo-label selection criteria to pick the suitable $(I_t, j)$ pairs for creating the target pseudo-label subset, $\mathcal{J}_\textit{inV}^t$. Here, $\mathcal{J}_\textit{outV}^t$ denotes the set of the $(I_t, j)$ pairs having \textit{joint-uncertainty} greater than a preset threshold $\alpha_h^\textit{th}$. %\pym{Please refer Supplementary for more details on the uncertainty thresholds.} 
Rest of the $(I_t, j)$ pairs can move either towards $\mathcal{J}_\textit{inV}^t$ or $\mathcal{J}_\textit{outV}^t$ over the course of adaptation training and are thus left untouched (no loss imposed).
\vspace{-2mm}
\begin{equation}
    \begin{aligned}
    \mathcal{J}_\textit{inV}^t = \{(I_t,j): \mathcal{H}(I_t,j) (|\tilde{q}_t^{(j)} - \mathcal{F}^{(j)}_q(\hat{q}_t^\prime)|)< \alpha_q^\textit{th}\}
    \end{aligned}
\end{equation}
\vspace{-2mm}

\noindent
The adaptation training involves minimizing $\boldsymbol{\mathcal{H}^{(t)}_{\mathcal{J}_\textit{inV}^t}}$ %(\ie joint-uncertainty for the samples in $\mathcal{J}_\textit{inV}^t$)
while maximizing $\boldsymbol{\mathcal{H}^{(t)}_{\mathcal{J}_\textit{outV}^t}}$ (\textit{joint-uncertainties} for $\mathcal{J}_\textit{inV}^t$ and $\mathcal{J}_\textit{outV}^t$ respectively), alongside minimizing the joint-supervision on target pseudo-labels, \ie $\boldsymbol{\mathcal{L}^\textit{OA}_\textit{pSup}}((I,j)\in \mathcal{J}_\textit{inV}^t)$

%%%%%%%%%%%%%%%%%%%%%%3DPW and 3DHP table %%%%%%%%%%%%%%%%%%%%%%
\begin{table*}[h!]
    \vspace{-3mm}
	\footnotesize
	\caption{ 
	Quantitative results on 3DPW and 3DHP. Numbers and layout taken from~\cite{Zhang_2021_ICCV}. $^*$ denotes inference stage (or online) optimization.
	\vspace{-6mm}
	}
	\centering
	\setlength\tabcolsep{4.0pt}
	\resizebox{0.99\textwidth}{!}{
	\begin{tabular}{ll|cc|cc|cc|cc}
% 	\toprule
    \hline
 		\multirow{2}{*}{\makecell{Adaptation \\ type}} & \multirow{2}{*}{Methods} & \multicolumn{2}{c|}{H3.6M$\rightarrow$3DPW} & 
 		 \multicolumn{2}{c|}{H3.6M$\rightarrow$3DHP} & 
 		 \multicolumn{2}{c|}{H3.6M$\rightarrow$SURREAL} &
 		 \multicolumn{2}{c}{H3.6M$\rightarrow$HumanEva}\\
 		\cline{3-10}
 		& & MPJPE$\downarrow$ & PA-MPJPE$\downarrow$ 
 		 & MPJPE$\downarrow$ & PA-MPJPE$\downarrow$ 
 		 & MPJPE$\downarrow$ & PA-MPJPE$\downarrow$ 
 		 & MPJPE$\downarrow$ & PA-MPJPE$\downarrow$\\
		\hline
		\hline
		\multirow{5}{*}{\makecell{General \\ Adaptation}} & DDC~\cite{Tzeng2014DeepDC} 
		 & 110.4 & 75.3 & 115.6 & 91.5 & 117.5 & 80.1 & 83.8 & 64.9\\
		 
		& DAN~\cite{long2015learning} 
		 & 107.5 & 73.2 & 109.5 & 89.2 & 114.2 & 78.4 & 78.5 & 62.7\\
		 
		& DANN~\cite{ganin2015unsupervised} 
		 & 106.3 & 71.1 & 107.9 & 88.0 & 113.6 & 77.2 & 76.3 & 60.8 \\
		 
		& Zhang \etal~\cite{Zhang_2021_ICCV} 
		 & 94.7 & 63.9 & 99.3 & 81.5 & 103.3 & 69.1 & 69.2 & 53.5\\
		
		& \textit{Ours} & \textbf{91.9} & \textbf{62.1} & \textbf{96.2} & \textbf{78.6} & \textbf{99.6} & \textbf{67.2} & \textbf{66.8} & \textbf{51.9}\\
		\hline 
		\multirow{3}{*}{\makecell{Test-time \\ Adaptation}} & ISO~\cite{inference_stage2020}$^*$ 
		 & - & 70.8 & - & 75.8 & - & - & - & -\\
		 
		& BOA~\cite{syguan2021boa}$^*$ 
		 & 92.1 & 58.8 & - & 77.4 & - & - & - & -\\
		 
		& \textit{Ours+ISO}$^*$ & \textbf{89.6} & \textbf{57.5} & \textbf{92.9} & \textbf{76.3} &    \textbf{96.4} & \textbf{65.1} & \textbf{65.2} & \textbf{50.1}\\
		\hline
	\end{tabular}}
	\vspace{-2mm}
	\label{tab:3dpw}
\end{table*} 
%%%%%%%%%%%%%%%%%%%%%%% 3DPW table from Zhang et al. ends  %%%%%%%%%%%%%%%%%%%%%%%%%%

\subsection{Inferring the final 3D pose}
In the proposed \textit{MRP-Net}, 3D pose can only be inferred through the ${B}_R$ branch (\ie $\hat{p}$). However, several recent works~\cite{moon2020i2l} advocate for a localization-based representation even for the 3D pose estimation by introducing another output for joint-wise depth-localization. Based on the pros and cons of both the modeling configurations, we decide to leverage the best of both worlds by training a fusion network to realize the final 3D pose prediction $\hat{p}^f$. The fusion network takes three inputs; a) 3D pose predictions via ${B}_R$ (\ie $\hat{p}$), b) 2D pose prediction via ${B}_L$ (\ie $\tilde{q}$), and c) the joint-confidences $\tilde{w}$. The fully-connected fusion network is trained to minimize a loss that is exactly similar to $\boldsymbol{\mathcal{L}_\textit{pSup}^{(t)}}$ but on samples from both source and target pseudo-label set. Please refer to Supplementary for more details. % on relative contributions of the inputs to the fusion network.

%%%%%%%%%%%%%%%%%%%%%%%%%%%%%%%%%%%%%%%%%%%%%%%%%%
%% https://www.askapache.com/online-tools/figlet-ascii/ (big) %%%
%  ______                      _                _       
% |  ____|                    (_)              | |      
% | |__  __  ___ __   ___ _ __ _ _ __ ___   ___| |_ ___ 
% |  __| \ \/ / '_ \ / _ \ '__| | '_ ` _ \ / _ \ __/ __|
% | |____ >  <| |_) |  __/ |  | | | | | | |  __/ |_\__ \
% |______/_/\_\ .__/ \___|_|  |_|_| |_| |_|\___|\__|___/
%             | |                                       
%             |_|                                
%%%%%%%%%%%%%%%%%%%%%%%%%%%%%%%%%%%%%%%%%%%%%%%%%%

\section{Experiments}
We demonstrate effectiveness of \textit{MRP-Net} (\textit{MRPN}) by evaluating it on a variety of cross-dataset settings.

\vspace{1mm}
\noindent
\textbf{Implementation details.}
We use \textit{ResNet-50}~\cite{he2016deep} (till Res-4f), pre-trained on the \textit{ImageNet}, as the common encoder $E$. %Inspired by I2L-MeshNet ~\cite{Moon_2020_ECCV_I2L-MeshNet}, 
The localization branch ${B}_L$ comprises of %consist of 3 blocks 
of transposed convolutional layers which progressively increase the spatial resolution to yield 17 heatmaps of size 56$\times$56.
The regression branch ${B}_R$ consists of a series of fully-connected (FC) layers which later bifurcate into two sub-branches to yield camera parameters $\hat{c}$ and local limb vectors $\hat{p}^l$. %Each sub-branch consists of three residual FC blocks. All FC layers use 1024 units and are followed by batch normalization and Leaky-ReLU non-linearity. 
We trained the framework on a NVIDIA P-100 GPU (16GB), with a batch size of eight. We employ separate Adam optimizers~\cite{kingma2014adam} for each loss term. See Suppl. for more details.

\vspace{1.5mm}
\noindent
\textbf{Datasets.} We use the following datasets.

\noindent

%\vspace{1mm}
\noindent
\textbf{a) SURREAL (S)} synthetic dataset~\cite{varol2017learning} is used both as source and target under different problem settings. Though the dataset encapsulates a wide range of diversity, synthetic-trained model suffers from poor generalization on natural images due to synthetic-to-real domain-shift. 
%  $\mathcal{D}_s$

%\vspace{1mm}
\noindent
\textbf{b) Backgrounds.} We use background images taken from; LSUN \cite{yu15lsun}, Google Street View \cite{6710175}, Natural Scenes \cite{10.1167/11.12.14}, and Campus Scenes \cite{Burge16849} to form the dataset, $\mathcal{D}_b$.

%\vspace{1mm}
\noindent
\textbf{c) Human3.6M (H).} For a fair evaluation, we use the standard, in-studio Human3.6M (H3.6M) dataset~\cite{ionescu2013human3} as either source or target domain in different problem settings.

%%%%%%%%%%%%%%%% The long table H3.6M %%%%%%%%%%%%%%%%%%%%%%%%%
\begin{table}[t!]
	%\footnotesize
	\caption{ 
	Quantitative comparison on Human3.6M. Our proposed method outperforms the prior-arts at various supervision levels. * denotes using MPII~\cite{andriluka14cvpr} with 2D pose annotations. \textsuperscript{+} denotes using additional in-the-wild data taken from the Internet. supervision-type on target (H3.6M) is indicated under the \textit{Supervision} column. \textit{Semi-sup (S1)} denotes 3D pose supervision only on subject S1. \vspace{-2mm}} % (row no. 1-3).
	\centering
	\setlength\tabcolsep{6.5pt}
	\resizebox{0.48\textwidth}{!}{
	\begin{tabular}{l|c|c|c}
% 	\begin{tabular}{ll|c|c}
    \hline
 		Methods & Supervision &
 		PA-MPJPE$\downarrow$ & MPJPE$\downarrow$ \\
		\hline\hline
% 		\rowcolor{grayLight}

		% 		\rowcolor{grayLight}
% 		Zhou \etal \cite{zhou2016sparse} & Full-3D & 
% 		106.7 \\
% 		%\rowcolor{grayDark}
% 		Chen~\etal~\cite{chen20173d} & Full-3D &	82.7 \\

		Martinez \etal \cite{martinez2017simple} & Full-3D & 52.5 & 67.5\\
		
% 		Li \etal \cite{Li_2020_CVPR} & Full-3D & 38.0\\
		
		Xu~\etal~\cite{Xu_2020_CVPR} & Full-3D & 36.2 & 45.6\\
		
		Chen~\etal~\cite{chen2020eccv} & Full-3D & 32.7 & 47.3 \\
		
	   % \pym{[R13] \etal} & Full-3D & 52.6 & 69.5 \\
		
% 		\rowcolor{grayLight}
		
		\hline

% 		\rowc7. & Wu \etal \cite{wu2016single} & Full-2D  & 98.4 \\
% 		%\rowcolor{grayDark}
%         \pym{[R17] \etal} & Full-2D & 113.6 & - \\
        
% 		Tung~\etal~\cite{tung2017adversarial} & Full-2D & 97.2 & - \\
		
% 		Chen~\etal~\cite{chen2019unsupervised} & Full-2D & 68.0 & -\\
		
% 		\pym{[R13] \etal }& {Full 2D} &
% 		66.0 & 91.8\\
		
% % 		%\rowcolor{grayDark}
% 		Wandt~\etal~\cite{wandt2019repnet} & Full-2D & 65.1 & 89.9\\
		
% 		Wang~\etal~\cite{wang_iccv19} & Full-2D &  
% 		62.8 & 86.4\\
		
% % 		 \rowcolor{gray!10}
% 		Kundu~\etal~\cite{kundu2020self}\textsuperscript{+} & Full-2D & 62.4 & -\\

% 		\textbf{\textit{Ours(S$\rightarrow$H, Weakly)}} & Full-2D & \textbf{59.2} & \textbf{77.2}\\
		
% 		\pym{[R14](MV train) \etal }& {Full 2D} &
% 		53.0 & 74.3\\
		
% 		\pym{[R37] \etal }& {Full 2D} &
% 		52.3 & 92.4\\
%         \hline\hline
        %\rowcolor{gray!10}
% 		Rhodin \etal \cite{rhodin2018unsupervised} & {Multi-view} & 98.2 & 131.7\\
		
% 		Kundu \etal \cite{kundu2020self} & {Multi-view} &		85.8 & -\\
		
% 		Iqbal \etal \cite{Iqbal_2020_CVPR}* & {Multi-view} & 55.9 & 69.1\\

% % 		\rowcolor{gray!10}
% 		\textbf{\textit{MRP-Net(PU)}} & {Multi-view} &		\textbf{53.7} & \textbf{65.9}\\
		
% 		\hline\hline
		
		Mitra~\etal~\cite{Mitra_2020_CVPR} & Semi-sup (S1) & 90.8 & 120.9\\
		
		Li~\etal~\cite{9010633}  & Semi-sup (S1) & 66.5 & 88.8\\
		
		Rhodin~\etal~\cite{rhodin2018unsupervised}  & Semi-sup (S1) & 65.1 & -\\
		
		Kocabas~\etal~\cite{kocabas2019self}  & Semi-sup (S1) & 60.2 & -\\

		Iqbal~\etal~\cite{Iqbal_2020_CVPR}* & Semi-sup (S1) & 51.4 & 62.8\\
		
% 		\rowcolor{gray!10}
		\textbf{\textit{Ours(S$\rightarrow$H, Semi)}} & Semi-sup (S1) & 
		\textbf{49.6} & \textbf{59.4}\\
		
		\hline
		
% 		\rowcolor{gray!10}
		Kundu~\etal~\cite{kundu2020self}\textsuperscript{+} & {No sup.} &
		99.2 & -\\ %\hline
		
		\textbf{\textit{Ours(S$\rightarrow$H)}} & No sup. & 
		\textbf{88.9} & \textbf{103.2}\\ %\hline
	
		\hline
	\end{tabular}}
	\vspace{-3mm}
	\label{tab:h36}
\end{table} 
%%%%%%%%%%%%%%%%%% The long table ends %%%%%%%%%%%%%%%%%%

%%%%%%%%%%%%%%%%% 3D pose results %%%%%%%%%%%%%%%%%
\begin{figure*}[h!]  %[!tbhp]%[!tbp][h!]
\begin{center}
	\vspace{-2mm}
    \includegraphics[width=0.98\linewidth]{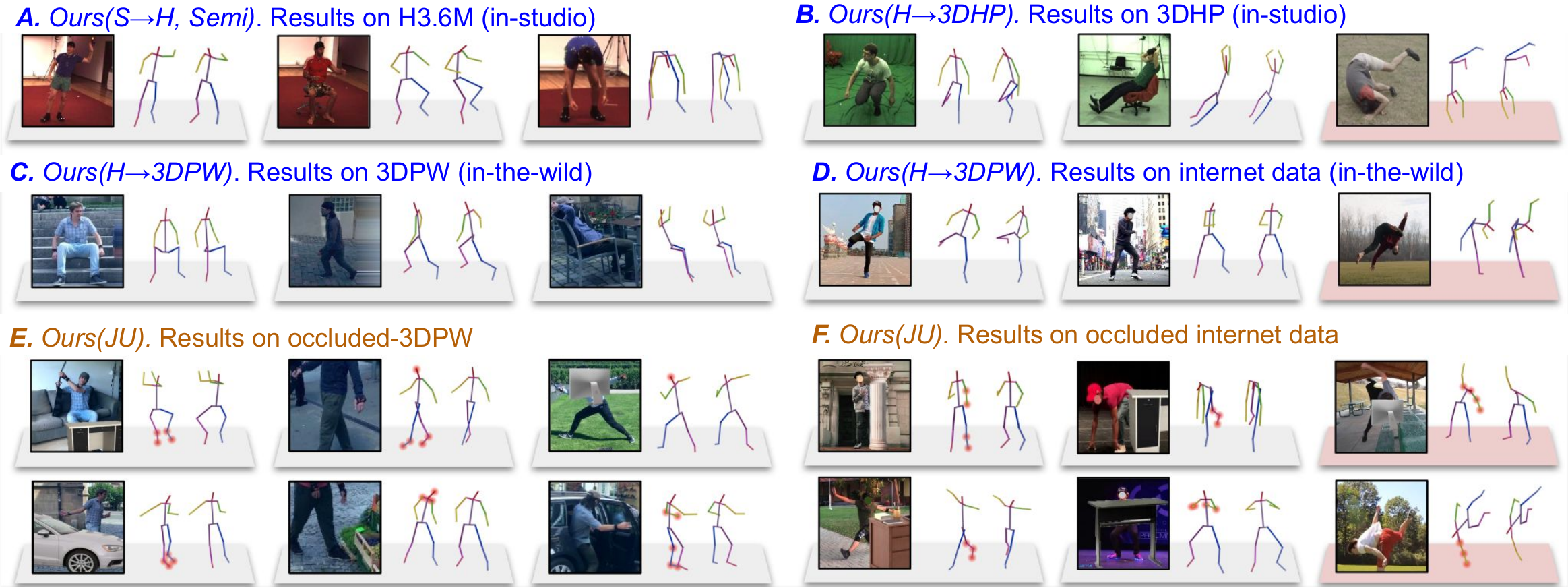}
    \vspace{-2mm}
	\caption{Qualitative analysis. 3D poses shown correspond to the original camera view and another azimuthal view %with azimuthal angle changed by
	(+30$^{\circ}$ or -30$^{\circ}$). %depending on best viewing angle. %%% For full-body images the results are obtained via our best model for the corresponding datasets. %We use \textit{Ours(SH$\rightarrow$ML)} to obtain pose prediction for the in-the-wild internet data. 
	For results in panel \textbf{E} and \textbf{F} %depicts results from \textit{Ours(JU)} where 
	the joints with uncertainty greater than a prefix threshold are highlighted with red-blobs. The model fails on rare poses, complex inter-limb occlusion and heavy background clutter as highlighted by red bases. Refer Suppl. for more results.
	%Qualitative analysis. We show results on four datasets - Human3.6M, 3DHP, 3DPW, and Web. \textit{M1-A} to \textit{M1-D} show results for \textit{MRP-Net(PU)}, \ie, our pose uncertainty network. \textit{M2-A} to \textit{M2-D} show results for \textit{MRP-Net(JU)}, \ie, our joint uncertainty network. Joint uncertainties are depicted by spheres around the occluded or truncated joints. \textit{MRP-Net} generalizes to both in-studio and in-the-wild %datasets. 
	}
    \vspace{-6mm}
    \label{fig:qualitative}  
\end{center}
\end{figure*}
%%%%%%%%%%%%%%%%%%%%%%%%%%%%%%%%%%%%%%%%%%%%%%%%%%%%%

\noindent
\textbf{d) Target datasets.} 3DPW~\cite{von2018recovering}, HumanEva~\cite{sigal2010humaneva},
%(\textbf{M}) and %LSP~\cite{johnson2010clustered} (\textbf{L}) 
and MPI-INF-3DHP (3DHP)~\cite{mehta2017monocular} are used as unlabeled target datasets to evaluate the unsupervised adaptation.

\noindent
\textbf{Occlusion simulation.}
We perform occlusion simulation on both source and target. The simulation process works on the corresponding full-body instances. We paste external objects (like cars, chair, wardrobe, etc.) to simulate occlusions, whereas truncation is simulated by randomly zooming into the top or bottom region of the full-body images.

\noindent
\textbf{Evaluation metrics.}
For a fair comparison, we evaluate our approach on the standard benchmark datasets described above. The standard mean per joint position error metric computed before and after Procrustes Alignment~\cite{gower1975generalized} are denoted as MPJPE and PA-MPJPE respectively~\cite{ionescu2013human3}.

\subsection{Quantitative analysis}

\noindent
\textbf{4.1.1. Synthetic to real adaptation.}
Labeled synthetic SURREAL (S) is used as the source domain, while the unlabeled target instances are taken from Human3.6M (H). Table~\ref{tab:h36} shows a comparison of \textit{Ours(S$\rightarrow$H)} on unsupervised and semi-supervised settings. Here, semi-supervised setting denotes 3D supervision only on subject S1. In spite of the huge domain shift between SURREAL and Human3.6M our adaptation strategy yields \textit{state-of-the-art} performance among the semi-supervised and unsupervised prior-arts.

\vspace{1mm}
\noindent
\textbf{4.1.2. In-studio to in-the-wild adaptation.}
We evaluate the performance of \textit{MRPN} in Table~\ref{tab:3dpw} on four target domains, i.e., 3DHP, 3DPW, SURREAL, and HumanEva datasets when using Human3.6M as the source domain. Our baseline is trained only on the source domain Human3.6M. A direct transfer on the target domains performs poorly attributed to the vast domain gap induced due to changing pose, appearance, and backgrounds between source and target datasets. Our pose-level adaptation strategy helps \textit{MRPN} improve upon the prior-arts even on in-the-wild 3DPW dataset by a significant margin, thereby validating ours superior generalizability. \textit{Ours+ISO} is the variant which uses ISO~\cite{inference_stage2020} for test time optimization.

\vspace{1mm}
\noindent
\textbf{4.1.3. Adaptation to partial body visibility.}
Table~\ref{tab:ablation} reports a quantitative analysis to highlight the merits of our design choices against the standard prior-art techniques. Under Joint-level adaptation, the baseline on row-1 shows transfer results on the target before adaptation. Row-2 baseline employs uncertainty maximization on target as well. Finally, row-3 uses target pseudo labels for self-training. \textit{Ours(JU)} outperforms LCR++~\cite{rogez2019lcr} even in the absence of 2D/3D supervision on in-the-wild datasets like MPII.

%%%%%%%%%%%%%%%% Ablation table %%%%%%%%%%%%%%%%%%
\begin{table}[t!]
    %\vspace{-1mm}
	%\footnotesize
	\caption{Ablation study. The column headings indicate usage of different loss terms during training. Ablations under \textit{pose-level} adaptation are evaluated on Human3.6M. * denotes inference without the fusion network. Ablations under \textit{joint-level} adaptation are evaluated on truncated/occluded 3DPW test-split, obtained via in-house occlusion simulations. Here, MPJPE is computed only for the true \textit{in-view} joints. B1 and B2 denotes our baselines under joint-level and pose-level adaptations respectively. 
	\vspace{-2mm}
	}
	\centering
	\setlength\tabcolsep{2pt}
	\resizebox{0.48\textwidth}{!}{
	\begin{tabular}{l|l|c|c|c|c}
	\hline

		%\hline
		%\cite{wandt2019repnet} & H36M & No sup. & \underline{81.8}  & \underline{54.8} & \underline{92.5}\\
% 		\rowcolor{gray!10} 
%%%%%%%%%%%%%%%%%%%%%%%%%%%%%%%%%%%%%%%%%%%%%%%%%%%%%%%%%%%
        
        \rowcolor{gray!10} \multicolumn{6}{c}{\texttt{Joint-level adaptation on {3DPW}}} \\ \hline
        
        No. & Method & \makecell{$\mathcal{L}_\textit{Sup}^\textit{OA} - \mathcal{H}^{(b)}_{\forall j}$}& $\mathcal{H}^{(t)}$ %_{\mathcal{J}_\textit{inV}^t}}$ 
        & $\mathcal{L}_\textit{pSup}^\textit{OA}$ & MPJPE$\downarrow$ \\ \hline \hline
        
		1. & \textit{B1(JU; H$\rightarrow$3DPW)}  & \cmark & - & - & 191.2 \\
		
		2. & \textit{B1(JU; H$\rightarrow$3DPW)} & \cmark & \cmark & - & 130.7 \\
	
        % \rowcolor{gray!10} 
% 		\textbf{\textit{MRP-Net(B) + DANN}} & SynD(3D) & 3DPW & 91.9 \\
		3. & \textit{Ours(JU; H$\rightarrow$3DPW)} & \cmark & \cmark & \cmark & \textbf{98.0} \\
		\cline{2-5}	
		4. & LCR-Net++~\cite{rogez2019lcr} & \multicolumn{3}{c|}{\makecell{H3.6M(3D), MPII(2D)}} & {104.9} \\
		\hline\hline
		%\textit{Ours(JU; SH$\rightarrow$ML)} & \multicolumn{2}{c|}{\makecell{SynD(3D), H3.6M(3D)}} & 3DHP(I) & - & \textbf{72.4} \\
		\rowcolor{gray!10} \multicolumn{6}{c}{\texttt{Pose-level adaptation on {Human3.6M}}}\\ \hline
	
 		No. & Method & \makecell{$\mathcal{L}_\textit{Sup}^{(s)} - \mathcal{U}^{(b)}$}& $\mathcal{U}^t$ & $\mathcal{L}_\textit{pSup}^{(t)}$ & MPJPE$\downarrow$ \\
		\hline \hline
		%\rowcolor{grayLight}
		%\cite{mehta2017monocular} & Full-3D & 76.5 & 40.8 & 117.6 \\
% 		\rowcolor{grayDark}
		5. & \textit{B2(S$\rightarrow$H)*+DANN}~\cite{ganin2016domain}
		& only $\mathcal{L}_\textit{Sup}^{(s)}$ & %\multicolumn{3}{c|}{\makecell{via domain classifier}}
		\multicolumn{2}{c|}{Standard DA} & {116.8} \\
		\cline{3-5}
		
        %6. & \textit{B2(S$\rightarrow$H)+DANN}~\cite{ganin2016domain} &  & \multicolumn{2}{c|}{} & {114.5} %\multicolumn{3}{c|}{\makecell{via domain classifier}}
		
		6. & \textit{B2(S$\rightarrow$H)}* & \cmark & - & - & {122.4} \\
		
% 		8. & \textit{B2(S$\rightarrow$H)} & \cmark & - & - & {122.1} \\

		7. & \textit{B2(S$\rightarrow$H)}* & \cmark & \cmark & - &  {113.4} \\
		
% 		10. & \textit{B2(S$\rightarrow$H)} & \cmark & \cmark & - &  {110.7} \\
		
		8. & \textit{Ours(S$\rightarrow$H)}* & \cmark & \cmark & \cmark & {106.3} \\
		
		9. & \textit{Ours(S$\rightarrow$H)} & \cmark & \cmark & \cmark & \textbf{103.2} \\
		\hline
	\end{tabular}}
	\vspace{-3mm}
	\label{tab:ablation}
\end{table}

%%%%%%%%%%%%%%%%% 3D pose results %%%%%%%%%%%%%%%%%
\begin{figure*}[h!]  %[!tbhp]%[!tbp][h!]
\begin{center}
% 	\includegraphics[width=0.98\linewidth]{image_final/aaai1_fig3_sid_final.pdf}
	%\vspace{-2mm}
    \includegraphics[width=0.98\linewidth]{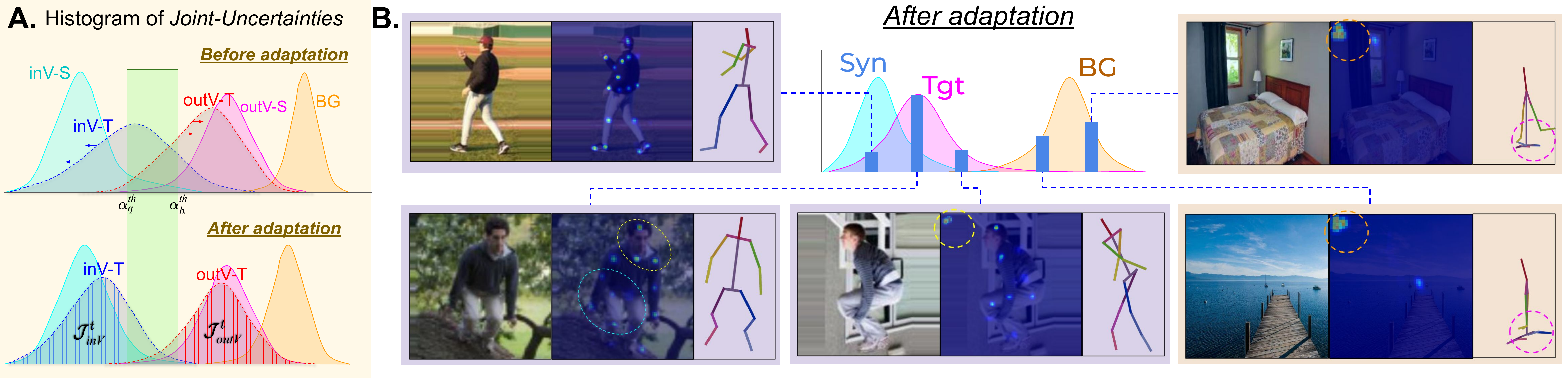}
    \vspace{-4mm}
	\caption{
        A. Histogram of samples from different domains (source, target, and background) along the joint uncertainty metric. B. Visualizing how the model choose to maximizes uncertainty for background (on right) and and occluded joints (on middle).
	}
    \vspace{-6mm}
    \label{fig:qualitative_mrc}  
\end{center}
\end{figure*}
%%%%%%%%%%%%%%%%%%%%%%%%%%%%%%%%%%%%%%%%%%%%%%%%%%%%%

%%%%%%%%%%%%%%%%% 3D pose results %%%%%%%%%%%%%%%%%
\begin{figure}[h!]  %[!tbhp]%[!tbp][h!]
\begin{center}
% 	\includegraphics[width=0.98\linewidth]{image_final/aaai1_fig3_sid_final.pdf}
%\vspace{-4mm}
    \includegraphics[width=0.95\linewidth]{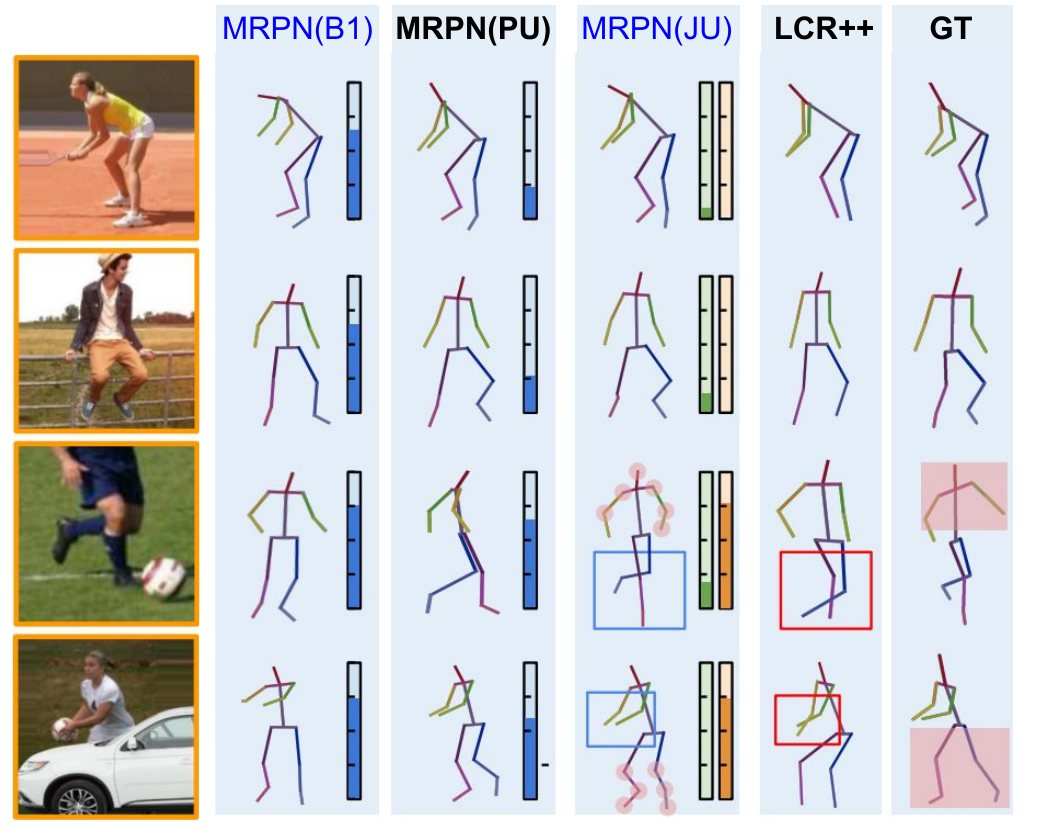}
	\vspace{-3mm}
	\caption{
	Pose prediction results showing measure of uncertainty values for pose, in-view, and out-view joints. The barometer height indicates high uncertainty. The blue, green and orange barometers indicate the prediction uncertainty for the full-pose, true-in-view joints and true-out-view joints respectively. %
	}
    \vspace{-8mm}
    \label{fig:qualitative_mrpn}  
\end{center}
\end{figure}
%%%%%%%%%%%%%%%%%%%%%%%%%%%%%%%%%%%%%%%%%%%%%%%%%%%%%

\vspace{1mm}
\noindent
\textbf{4.1.4. Ablation study.}
In Table~\ref{tab:ablation}, under \textit{pose-level} adaptation, the baseline on row-5 uses \textit{MRPN} architecture while employing an adversarial discriminator based discrepancy minimization on encoder features (DANN). The baseline on row-6 shows transfer results on the target before adaptation. Row-8 baseline employees self-training unlike in row-7 where only the target uncertainty is minimized. Our final model is depicted in row-9 where the pose predictions are obtained via the fusion-network unlike in row-4.

\subsection{Qualitative evaluation and limitations}
Fig.~\ref{fig:qualitative} and Fig.~\ref{fig:qualitative_mrpn} analyze our pose prediction results across variations in pose complexity, occlusion/truncation scenarios and environmental conditions (\ie in-studio and in-the-wild). Fig.~\ref{fig:qualitative} presents extensive in-the-wild and partial body visibility scenarios. We also show 
%3D pose estimation 
results on images randomly taken from online sources. \textit{MRP-Net} successfully
estimates reasonable 3D poses for most of the occluded and truncated cases. However, our model may fail under certain drastic scenarios such as multi-level body-part occlusion, high background clutter, and rare athletic poses. Fig.~\ref{fig:qualitative_mrpn} gives an insight into how \textit{MRP-Net} adapts to both unoccluded and partial-body visibility scenarios with predictions better than LCR++~\cite{rogez2019lcr}.
\textit{MRPN(B1)} indicates the occlusion-aware network before the adaptation training. Without adaptation on target samples, the model predicts with high uncertainty. \textit{MRPN(PU)} and \textit{MRPN(JU)} indicate the final networks after the \textit{pose-level} and \textit{joint-level} adaptations. %respectively. 
\textit{MRPN(PU)} is not tuned to work on occluded/truncated images and thus yields a higher uncertainty for the bottom two images. Whereas, the uncertainty predictions of \textit{MRPN(JU)} for the green and orange barometer yield the expected behaviour. The red-blocks under GT column segregate the true \textit{out-view} joints. The in-view joint predictions of \textit{MRPN(JU)} match with the same under GT. 

\vspace{1mm}
\noindent
\textbf{Societal impacts.} We do not foresee a direct negative societal impact from our framework. However, it may be leveraged for human-tracking applications. We urge the readers to make ethical and responsible use of our work. 

\vspace{1mm}
\noindent
\textbf{4.2.1. Model interpretability.}
We also perform a thorough qualitative study to interpret the behaviour of our network for a wide variety of in-distribution and out-of-distribution samples. In Fig.~\ref{fig:qualitative_mrc}, we analyze how samples from different domains (such as source, target, and backgrounds) are distributed along the uncertainty metrics, \ie the \textit{pose-uncertainty} and \textit{joint-uncertainty}. This gives an insight into how \textit{MRPN} caters to OOD samples as well as the uncertainty associated with partial body visibility. Fig.~\ref{fig:qualitative_mrc}\redcolor{A} shows histogram of the predicted \textit{joint-uncertainties} for the true \textit{in-view} and \textit{out-view} joints separately for source (\ie \texttt{inV-S} and \texttt{outV-S}) and target (\ie \texttt{inV-T} and \texttt{outV-T}). \texttt{BG} denotes the histogram of all \textit{out-view} joints for backgrounds. The shaded regions in the bottom panel depicts $\mathcal{J}_\textit{inV}^t$ and $\mathcal{J}_\textit{outV}^t$ which are segregated using the preset thresholds $\alpha_q^\textit{th}$ and $\alpha_h^\textit{th}$ respectively (edges of the green-box). Our adaptation algorithm succeeds to separate \textit{inV-T} and \textit{outV-T} over the course of adaptation training. 
Next, Fig.~\ref{fig:qualitative_mrc}\redcolor{B} shows a similar analysis for \textit{pose-uncertainties}. We show five different examples sampled from different regions of the histogram-bins. %%
In the right-panel, we show that to maximize \textit{pose-uncertainty} for backgrounds (OOD samples), \textit{MRPN} estimates the 2D landmarks and 3D pose points separated towards opposite diagonal corners. Here, the 2D landmarks are collapsed to the top-left corner whereas the root joint (pelvis) of the model-based 3D predictions are seemed to have collapsed towards the bottom-right corner. %%%
In the bottom-panel, for uncertain target instances, we see two peaks in the joint heatmap distributions; one at the top-left corner (OOD-related) and the other near the actual joint location. During adaptation, the OOD-related peak suppresses while the joint-related peak rises to simultaneously reduce the uncertainty while converging towards the true pose outcome. %%%
Finally, on the left panel, \textit{joint-level} uncertainty is indicated by the entropy of heatmap distribution.

% ***************sid*********
%\vspace{-2mm}
\section{Conclusion}\vspace{-1mm}

We presented a multi-representation pose network that embraces the pros and cons of both model-free and model-based pose representations to realize a disagreement based pose-uncertainty measure. We develop learning techniques to make the model behave differently for the in-domain and out-of-domain scenarios. %at both pose and joint level granularity. 
Later, the same instigated behaviour is used to devise effective unsupervised adaptation objectives. Formalizing prediction uncertainty in the presence of temporal context remains to be explored in future.

\vspace{1mm}
\noindent \textbf{Acknowledgements.} This work is supported by %a Google PhD Fellowship (Jogendra) and a grant from 
Uchhatar Avishkar Yojana (UAY, IISC\_010), MoE, Govt. of India.

%% \pagebreak

\appendix

%%%%%%%%%%%%%%%%%%%%%%%%%%%%%%%%%%%%%%%%%%%%%%%%%%%%%%
%%%%%%%%%%%%%%%%%%%%%%%%%%%%%%%%%%%%%%%%%%%%%%%%%%%%%%
% \begin{multicols*}{2}
% \vspace{5mm}
% \vspace{2mm}
% \noindent \textbf{\Large Supplementary Material}
% \vspace{2mm}
% \end{multicols*}

%%%%%%%%%%%%%%%%% Fig Arch %%%%%%%%%%%%%%%%%
\begin{figure*}[!h] %[!tbp]%[!b]%[!tbp][h!]
\begin{center}
    \vspace{8mm}
	\includegraphics[width=0.98\linewidth]{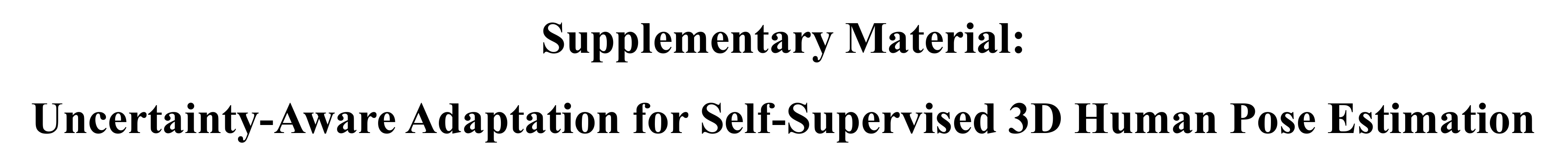}  
    \vspace{9mm}
\end{center}
\end{figure*}
%%%%%%%%%%%%%%%%% Fig Arch ends %%%%%%%%%%%%%%%%%

\noindent The supplementary document is organized as follows: 
%\vspace{-1mm}
\begin{itemize}
%\item Section ~\ref{sec:1}: Data preparation
%\vspace{1mm}
\item Section ~\ref{sec:2}: Notations
%\vspace{1mm}
\item Section ~\ref{sec:3}: Training algorithms 
%\vspace{1mm}
% \item Section ~\ref{sec:4}: Transforming local vectors to canonical pose
%\vspace{1mm}
\item Section ~\ref{sec:5}: Network architecture
%\vspace{1mm}
\item Section ~\ref{sec:7}: Qualitative analysis
%\item Section \ref{sec:sec_limitations}: Limitations of the proposed framework
\end{itemize}

\section{Notations}
\label{sec:2}
Most of the notations used in this paper are summarized in Table~\ref{tab:notation_table}. In the first part, we list the general architecture related notations. Next, we group other notations into a) output of $B_L$, b) output of $B_R$, c) datasets, and finally the adaptation training related notations for both d) pose-level and e) joint-level adaptation.

%%%%%%%%%%%%%%% Table starts $$$$$$$$$$$$$$$$$$$$$$$$$$$
\begin{table}[!th]
%\vspace{-2mm}
    \centering
    \renewcommand{\arraystretch}{1.17}
    \setlength{\tabcolsep}{0.4mm}
    \caption{Notation Table}
    \label{tab:notation_table}
    %\vspace{-1mm}
    \resizebox{1\linewidth}{!}{
        \begin{tabular}{|c|c|l|}
            \hline
            & \textbf{Symbol} & \textbf{Description} \\
            
            \hline \hline
            \multirow{6}{*}{{\begin{turn}{90}{\small Miscellaneous}\end{turn}}} 
            & $J$ & Total no. of joints (17) indexed by $j$ \\
            & $E$ & Encoder as the common backbone CNN \\
            & $B_L$ & Localization branch (outputs heatmaps) \\
            & $B_R$ & Regression branch (outputs 3D pose) \\
            & $T_\textit{FK}$ & Forward-kinematics operation \\
            & $T_c$ & Weak-perspective projection operation \\
            
            \hline \hline
            \multirow{3}{*}{{\begin{turn}{90}{\small o/p of $B_L$}\end{turn}}} 
            & $\tilde{h}^{(j)}$ & Heatmap PDF for $j^\textit{th}$ joint $\in \mathbb{R}^{H^\prime\times W^\prime}$ \\
            & $\tilde{q}^{(j)}$ & 2D pose coordinates for the $j^\textit{th}$ joint $\in \mathbb{R}^2$ \\
            & $\tilde{w}^{(j)}$ & Joint confidences for the $j^\textit{th}$ joint $\in [0,1]$ \\
            
            \hline \hline
            \multirow{5}{*}{{\begin{turn}{90}{\small O/p of $B_R$}\end{turn}}} 
            & $\hat{p}^l$ & Local pose vectors (parent-relative) $\in\mathbb{R}^{J\times 3}$ \\
            & $\hat{c}$ & Camera parameters (3 angles, 1 scale, 2 translations) \\
            & $\hat{p}^c$ & Canonical 3D pose coordinates $\in\mathbb{R}^{J\times 3}$ \\
            & $\hat{p}$ & Camera-relative 3D pose coordinates $\in\mathbb{R}^{J\times 3}$ \\
            & $\hat{q}$ & projected 2D pose coordinates $\in\mathbb{R}^{J\times 2}$ \\
            
            \hline \hline 
            \multirow{3}{*}{{\begin{turn}{90}{\small Datasets}\end{turn}}} 
            & $\mathcal{D}_s$, $\mathcal{D}_t$ & Labeled source and unlabeled target datasets (full-body) \\
            & $\mathcal{D}_s^O$, $\mathcal{D}_t^O$ & Source and target datasets with occlusion/truncation \\
            & $\mathcal{D}_b$ & A dataset of background images (other than human) \\
            
            \hline \hline 
            \multirow{7}{*}{{\begin{turn}{90}{\small Pose-level}\end{turn}}} 
            & $\mathcal{U}(I)$ & Pose-level uncertainty for a given image \\
            & $\mathcal{L}_\textit{Sup}^{(s)}$ & Supervised loss on $\mathcal{D}_s$ samples (minimized) \\
            & $\mathcal{U}^{(s)}$ & Pose-uncertainty of $\mathcal{D}_s$ samples (minimized) \\
            & $\mathcal{U}^{(b)}$ & Pose-uncertainty of $\mathcal{D}_b$ samples (maximized) \\
            & $\mathcal{U}^{(t)}$ & Pose-uncertainty of $\mathcal{D}_t$ samples (minimized) \\
            & $\mathcal{L}_\textit{pSup}^{(t)}$ & Loss on pseudo-label target subset $\mathcal{D}_t^\textit{pl}$ (minimized) \\
            & $ \alpha_p^\textit{th}$ & Threshold to select pseudo-labeled target subset $\mathcal{D}_t^\textit{pl}$  \\
            
            \hline \hline 
            \multirow{9}{*}{{\begin{turn}{90}{\small Joint-level}\end{turn}}}
             & $\mathcal{H}(I,j)$ & Joint-level uncertainty (JU) for a given image, joint-id pair \\ \vspace{0.4mm}
            & $\mathcal{L}_\textit{Sup}^\textit{OA}$ & Occlusion-aware supervised loss on $\mathcal{D}_s^O$ (minimized) \\ \vspace{0.4mm}
            & $\mathcal{H}^{(s)}_{\mathcal{J}^s_\textit{outV}}$ & JU of true out-view joints of $\mathcal{D}_s^O$ (maximized) \\ \vspace{0.4mm}
            & $\mathcal{H}^{(b)}_{\forall j}$ & JU of all joints for backgrounds $\mathcal{D}_b$ (maximized) \\ \vspace{0.4mm}
            & $\mathcal{H}^{(t)}_{\mathcal{J}^t_\textit{inV}}$ & JU of pseudo-selected in-view joints of $\mathcal{D}_t^O$ (minimized) \\ \vspace{0.4mm}
            & $\mathcal{H}^{(t)}_{\mathcal{J}^t_\textit{outV}}$ & JU of pseudo-selected out-view joints of $\mathcal{D}_t^O$ (maximized) \\ \vspace{0.4mm}
            & $\mathcal{L}_\textit{pSup}^\textit{OA}$ & Loss on pseudo-labeled target set $(I,j)\in \mathcal{J}_\textit{inV}^t$ (minimized) \\ 
            & $ \alpha_q^\textit{th}$ & Threshold to select pseudo-labeled target in-view set $\mathcal{J}^t_\textit{inV}$  \\
            & $ \alpha_h^\textit{th}$ & Threshold to select pseudo-labeled target out-view set $\mathcal{J}^t_\textit{outV}$  \\
            
            \hline
        \end{tabular}
        \vspace{-10mm}
    }
\end{table}
%%%%%%%%%%%%%%%%%% Algo-1 starts %%%%%%%%%%%%%%%%%%%%%%
\begin{algorithm}[t]
\caption{Training algorithm for pose-level adaptation.}
\label{algo:algo1}
\begin{algorithmic}[1]
\State \textbf{Input:} Labeled source dataset $\mathcal{D}_s$, unlabeled target dataset $\mathcal{D}_t$, and the background dataset $\mathcal{D}_b$. Let $\Theta$ denote the learnable parameters of the \textit{MRP-Net} architecture (excluding the fusion network).

\vspace{1mm}
\State \textbf{while} $\textit{iter}<\textit{MaxIter}$ \textbf{do}
\vspace{1mm}
%%%%%%%%%%%%%%%%%%%%%%%%%%%%%%%%%%%%%%%%%%%%
\Statex \textit{\underline{\textbf{A. Pseudo-label update} {(after each $K_\textit{interval}$)}.}}
\If{$\textit{iter} \pmod{K_\textit{interval}}=0$}
\State \textbf{Compute} $\mathcal{D}_t^\textit{pl}$ where $\hat{q}_t$ and $\tilde{q}_t^\prime$ are obtained using 
\Statex \hspace{4mm} current state of network parameters $\Theta$, as follows:
\Statex \hspace{4mm} $\mathcal{D}_t^\textit{pl} = \{I_t: (|\hat{q}_t - \mathcal{F}_q(\tilde{q}_t^\prime)| + |\tilde{q}_t - \mathcal{F}_q(\hat{q}_t^\prime)|)< \alpha_p^\textit{th}\}$

\EndIf
%%%%%%%%%%%%%%%%%%%%%%%%%%%%%%%%%%%%%%%%%%%%

\vspace{1mm}
\Statex \textit{\underline{\textbf{B. Adaptation training} \textit{(for pose-level adaptation)}.}}
\vspace{1mm}
\State \textbf{Update} $\Theta$ by minimizing $\mathcal{L}_h(\hat{h}, h_\textit{gt})$, $\mathcal{L}_p(\hat{p}, p_\textit{gt})$, and $\mathcal{U}^{(s)}$ (\ie the first two terms under $\mathcal{L}_\textit{Sup}^{(s)}$) on a mini-batch of $\mathcal{D}_s$ using separate Adam optimizers.

\vspace{1mm}
\State \textbf{Update} $\Theta$ by maximizing $\mathcal{U}^{(b)}$ on a mini-batch of $\mathcal{D}_b$ using Adam optimizer.

\vspace{1mm} %s on pseudo-label 
\State \textbf{Update} $\Theta$ by minimizing $\mathcal{U}^{(t)}$ on a mini-batch of $\mathcal{D}_t$ using Adam optimizer.

\vspace{1mm}
\State \textbf{Update} $\Theta$ by maximizing $\sum_j \tilde{w}^{(j)}\mathcal{L}^{(j)}(\tilde{h}, h_\textit{gt}^\textit{pl})$ and $\sum_j \tilde{w}^{(j)}\mathcal{L}^{(j)}(\hat{p}, p_\textit{gt}^\textit{pl})$ (\ie the two terms under $\mathcal{L}_\textit{pSup}^{(t)}$) using separate Adam optimizers.

\State \textbf{end while}
\end{algorithmic}
\end{algorithm}
%%%%%%%%%%%%%%%%%% Algo-1 ends %%%%%%%%%%%%%%%%%%%%%%

%%%%%%%%%%%%%%%%%%%%%%%%%%%%%%%%%%%%%%%%%%%%%%%%%%%%%%%%%%%%%%%%%%%%%%%%%%%%

%%%%%%%%%%%%%%%%%% Algo-3 starts %%%%%%%%%%%%%%%%%%%%%%
\begin{algorithm}[b]
\caption{Training algorithm for the fusion network.}
\label{algo:algo3}
\begin{algorithmic}[1]
\State \textbf{Input:} Labeled source dataset $\mathcal{D}_s$ and the pseudo-labeled target subset $\mathcal{D}_t^\textit{pl}$. The network takes 3 inputs: a) 3D pose predictions via ${B}_R$ (\ie $\hat{p}$), b) 2D pose prediction via ${B}_L$ (\ie $\tilde{q}$), and c) the joint-confidences $\tilde{w}$ via $B_L$. Let $\theta^f$ denote the learnable parameters of the fusion network.

\vspace{1mm}
\While{$\textit{iter}<\textit{MaxIter}$}

\vspace{1mm}
\State \textbf{Update} $\theta^f$ to minimize $\mathcal{L}_p(\hat{p}^f, p_\textit{gt})$ on a mini-batch 
\Statex \hspace{4mm} of $\mathcal{D}_s$ using Adam optimizer.

\vspace{1mm}
\State \textbf{Update} $\theta^f$ to minimize $\sum_{j=1}^J \tilde{w}^{(j)}\mathcal{L}^{(j)}(\hat{p}^f, p_\textit{gt}^\textit{pl})$ on
\Statex \hspace{4mm} a mini-batch of $\mathcal{D}_t^\textit{pl}$ using Adam optimizer.

\EndWhile
\end{algorithmic}
\end{algorithm}
%%%%%%%%%%%%%%%%%% Algo-3 ends %%%%%%%%%%%%%%%%%%%%%%

%%%%%%%%%%%%%%%%%%%%%%%%%%%%%%%%%%%%%%%%%%%%%%%%%%%%%%%%%%%%%%%%%%%%%%%%%%%%
%%%%%%%%%%%%%%%%%%%%%%%%%%%%%%%%%%%%%%%%%%%%%%%%%%%%%%%%%%%%%%%%%%%%%%%%%%%%

%%%%%%%%%%%%%%%%%% Algo-2 starts %%%%%%%%%%%%%%%%%%%%%%
\begin{algorithm}[!t]
\caption{Training algorithm for joint-level adaptation.}
\label{algo:algo2}
\begin{algorithmic}[1]
\State \textbf{Input:} Labeled source dataset $\mathcal{D}^O_s$, unlabeled target dataset $\mathcal{D}^O_t$, and the background dataset $\mathcal{D}_b$. Let $\Theta$ denote the learnable parameters of the \textit{MRP-Net} architecture (excluding the fusion network).

\vspace{1mm}
\State \textbf{while} $\textit{iter}<\textit{MaxIter}$ \textbf{do}
\vspace{1mm}
%%%%%%%%%%%%%%%%%%%%%%%%%%%%%%%%%%%%%%%%%%%%
\Statex \textit{\underline{\textbf{A. Pseudo-label update} {(after each $K_\textit{interval}$)}.}}
\If{$\textit{iter} \pmod{K_\textit{interval}}=0$}
\State \textbf{Compute} $\mathcal{J}_\textit{inV}^t$ and $\mathcal{J}_\textit{outV}^t$, where $\hat{q}_t$ and $\tilde{q}_t^\prime$ are 
\Statex \hspace{3.5mm} obtained using the current state of the network 
\Statex \hspace{3.5mm} parameters $\Theta$, as follows:

\Statex \hspace{3.5mm} $\mathcal{J}_\textit{inV}^t = \{(I_t,j): \mathcal{H}(I_t,j) (|\tilde{q}_t^{(j)} - \mathcal{F}^{(j)}_q(\hat{q}_t^\prime)|)< \alpha_q^\textit{th}\}$
\Statex \hspace{3.5mm} $\mathcal{J}_\textit{outV}^t = \{(I_t,j): \mathcal{H}(I_t,j) (|\tilde{q}_t^{(j)} - \mathcal{F}^{(j)}_q(\hat{q}_t^\prime)|)> \alpha_h^\textit{th}\}$

\EndIf
%%%%%%%%%%%%%%%%%%%%%%%%%%%%%%%%%%%%%%%%%%%%

\vspace{1mm}
\Statex \textit{\underline{\textbf{B. Adaptation training} \textit{(for joint-level adaptation)}.}}
\vspace{1mm}
\State \textbf{Update} $\Theta$ by minimizing $\mathbbm{1}_{(I,j)\in\mathcal{J}_\textit{inV}^s} \mathcal{L}_h^{(j)}(\tilde{h}, h_\textit{gt})$ and $\mathbbm{1}_{(I,j)\in\mathcal{J}_\textit{inV}^s} \mathcal{L}_p^{(j)}(\hat{p}, p_\textit{gt})$ (\ie the first 2 terms under $\mathcal{L}_\textit{Sup}^\textit{OA}$) on a mini-batch of $\mathcal{D}_s^O$ using separate Adam optimizers.

\vspace{1mm}
\State \textbf{Update} $\Theta$ to maximize $\mathcal{H}^{(s)}_{\mathcal{J}^{s}_\textit{outV}} = \mathbbm{1}_{(I,j)\in\mathcal{J}_\textit{outV}^s} \mathcal{H}(I,j)$ on a mini-batch of $\mathcal{D}_s^O$ using Adam optimizer.

\vspace{1mm}
\State \textbf{Update} $\Theta$ to maximize $\mathcal{H}^{(b)}_{\forall j}$ on a mini-batch of $\mathcal{D}_b$ using Adam optimizer.

\vspace{1mm}
\State \textbf{Update} $\Theta$ to minimize $\mathcal{H}^{(t)}_{\mathcal{J}_\textit{inV}^t} = \mathbbm{1}_{(I,j)\in\mathcal{J}_\textit{inV}^t} \mathcal{H}(I,j)$ on a mini-batch of $\mathcal{D}_t^O$ using Adam optimizer.

\vspace{1mm}
\State \textbf{Update} $\Theta$ to maximize $\mathcal{H}^{(t)}_{\mathcal{J}_\textit{outV}^t} = \mathbbm{1}_{(I,j)\in\mathcal{J}_\textit{outV}^t} \mathcal{H}(I,j)$ on a mini-batch of $\mathcal{D}_t^O$ using Adam optimizer.

\vspace{1mm}
\State \textbf{Update} $\Theta$ to maximize $\sum_{j\in\mathcal{J}_\textit{inV}^t} \tilde{w}^{(j)}\mathcal{L}^{(j)}(\tilde{h}, h_\textit{gt}^\textit{pl})$ and $\sum_{j\in\mathcal{J}_\textit{inV}^t} \tilde{w}^{(j)}\mathcal{L}^{(j)}(\hat{p}, p_\textit{gt}^\textit{pl})$ (\ie the two terms under $\mathcal{L}_\textit{pSup}^\textit{OA}$) using separate Adam optimizers. Here, $\tilde{w}^{(j)}$ is normalized such that $\sum_{j\in\mathcal{J}_\textit{inV}^t} \tilde{w}^{(j)}=1$.

\State \textbf{end while}
\end{algorithmic}
\end{algorithm}
%%%%%%%%%%%%%%%%%% Algo-2 ends %%%%%%%%%%%%%%%%%%%%%%

%%%%%%%%%%%%%%%%%%%%%%%%%%%%%%%%%%%%%%%%%%%%%%%%%%%%%%%%%%%%%%%%%%%%%%%%%%%%
%%%%%%%%%%%%%%%%%%%%%%%%%%%%%%%%%%%%%%%%%%%%%%%%%%%%%%%%%%%%%%%%%%%%%%%%%%%%

%%%%%%%%%%%%%%%%%%%%%%%%%%%%%%%%%%%%%%%%%%%%%%%%%%%%%%%%%%%%%%%%%%%%%%%%%%%%

\section{Training algorithms}
\label{sec:3}
In this section, we clearly discuss the training algorithms which could not be included in the main paper. Algo.~\ref{algo:algo1} and Algo.~\ref{algo:algo2} show the training algorithm for pose-level and joint-level adaptation respectively. We simultaneously train on samples from all the three datasets, \ie on $\mathcal{D}_s$, $\mathcal{D}_t$, and $\mathcal{D}_b$ for pose-level adaptation and on $\mathcal{D}^O_s$, $\mathcal{D}^O_t$, and $\mathcal{D}_b$ for joint-level adaptation. The pseudo-label selection procedure is clearly explained in both the algorithms (refer Table~\ref{tab:notation_table} for a description of the notations). Though we use the above for pose-level adaptation for a fair prior-art benchmarking, one %can always 
is always free to 
relax this assumption. \textbf{a)} Under pose-level DA, synthetic training on $\mathcal{D}_s^O$ (truncated+full) would make it applicable for both full and truncated target. In Fig. {\color{red}5}, notice the medium level uncertainty elicited by \texttt{MRPN(PU)} for truncated target (a desirable behaviour). \textbf{b)} On the other hand, joint-level adaptation already suits to both the scenarios (\texttt{MRPN(JU)} in Fig.~\ref{fig:fig4}

Algo.~\ref{algo:algo3} shows a detailed training procedure to prepare the fusion network for the pose-level adaptation scenario. We prepare a separate fusion network for the joint-level adaptation. {Table~\ref{tab:fusion} reports
relative contributions of $B_R$ and $B_L$ outputs against the fused.} In case of joint level adaptation the loss-term in {\color{red}L3} of Algo.~\ref{algo:algo3} is replaced by $\mathbbm{1}_{(I,j)\in\mathcal{J}_\textit{inV}^s} \mathcal{L}_p^{(j)}(\hat{p}, p_\textit{gt})$ (the second loss-term in {\color{red}L6} of Algo.~\ref{algo:algo2}). Similarly, the loss-term in {\color{red}L4} of Algo.~\ref{algo:algo3} is replaced by $\sum_{j\in\mathcal{J}_\textit{inV}^t} \tilde{w}^{(j)}\mathcal{L}^{(j)}(\hat{p}, p_\textit{gt}^\textit{pl})$ (the second loss-term in {\color{red}L11} of Algo.~\ref{algo:algo2}).

We trained the framework on an NVIDIA P-100 GPU (16GB) with a batch size of 8. We employ separate Adam optimizers~\cite{kingma2014adam} for each loss term. Please refer Fig~\ref{fig:hyperparam_plot} for sensitivity analysis. Note that, we use fixed threshold values across all adaptation settings in Sec {\color{red}4.1}.

\noindent
\textbf{Importance of OOD images.} We would like to reiterate that the background images represent an objective segregation of hard-OOD samples. The poses outside of the training distribution are critical to identify and we segregate them via the pseudo-label subset selection criteria (Eq. {\color{red}4}). Eq. {\color{red}5} selectively imposes a strong loss on the more confident target samples. It is to be noted that, such segregation is highly subjective, and treating these soft-OOD samples as hard-OOD deteriorates the generalization performance.

\begin{table*}[t]
%\vspace{-7mm}
\begin{minipage}[t]{0.38\textwidth}
%\begin{table}[]
\addtolength{\tabcolsep}{2pt}
    \centering
    \caption{Assets and the corresponding Licenses \vspace{-3.7mm}}
    \label{tab:license}
    \resizebox{\linewidth}{!}{%
    \begin{tabular}{|l|l|}
         \hline
         Asset used & License\\
         \hline
         \hline
         %\hline
         Human3.6M \href{http://vision.imar.ro/human3.6m/description.php}{\faExternalLink}~\cite{ionescu2013human3} & Limited license for academic use~\href{http://vision.imar.ro/human3.6m/eula.php}{\faExternalLink} \\
         %\hline
         MPI-INF-3DHP~\href{http://gvv.mpi-inf.mpg.de/3dhp-dataset/}{\faExternalLink}~\cite{mehta2017monocular} & Limited license for academic use~\href{http://gvv.mpi-inf.mpg.de/3dhp-dataset/}{\faExternalLink} \\
         3DPW~\href{https://virtualhumans.mpi-inf.mpg.de/3DPW/}{\faExternalLink}~\cite{von2018recovering} & Limited license for academic use~\href{https://virtualhumans.mpi-inf.mpg.de/3DPW/license.html}{\faExternalLink} \\
         HumanEva~\href{http://humaneva.is.tue.mpg.de/}{\faExternalLink}~\cite{sigal2010humaneva} & Limited license for academic use~\href{http://humaneva.is.tue.mpg.de/data_license}{\faExternalLink} \\
         %\hline
         %\hline
         SURREAL~\href{https://www.di.ens.fr/willow/research/surreal/data/}{\faExternalLink}~\cite{varol2017learning}& Limited license for academic use~\href{https://www.di.ens.fr/willow/research/surreal/data/license.html}{\faExternalLink} \\
         \hline
        %  Sports-1M~\href{https://cs.stanford.edu/people/karpathy/deepvideo/}{\faExternalLink}~\cite{KarpathyCVPR14} & CC BY 3.0~\href{https://cs.stanford.edu/people/karpathy/deepvideo/}{\faExternalLink} \\
        %  \hline
    \end{tabular}}
%\end{table}
\end{minipage}
\hfill
\begin{minipage}[t]{0.25\textwidth}
%%%%%%%%%%%%%%%% Complexity table %%%%%%%%%%%%%%%%%%
%\begin{table}[t!]
    %\vspace{-1mm}
% 	\footnotesize
% 	\captionof{table}{\small
% 	Model complexity
% 	\vspace{-3mm}
% 	}
% 	\centering
% 	\setlength\tabcolsep{0.5pt}
% 	\resizebox{\textwidth}{!}{
% 	\begin{tabular}{l|c|c}
% 	%\toprule
%  		Method & Inference & \#Params  \\
% 		\hline\hline

% 	    SH + PoseNet\cite{} & 95 ms & (25+25)M \\

% 	    %& + 20 ms (2D to 3D) & + Filler(2 layers) + 0.165M (SJA)\\
		
% 		\textit{Ours(S$\rightarrow$ML)} & 32 ms & 48M \\
% 		%& [root relative 3D]  & (+100.7M without GAP for 3D branch)\\
%         %& 35-37 ms [3D + heatmaps] & \\
% 		%\hline

% 		\bottomrule
% 	\end{tabular}}
% 	%\vspace{-2mm}
% 	\label{tab:complexity}
%\end{table} 
%%%%%%%%%%%%%%%%%%%%%%% Complexity table ends  %%%%%%%%%%%%%%%%%%%%%%%%%%%%%%%%%%%%%%%%%%%%%%
%%%%%%%%%%%%%%%% Fusion Network table %%%%%%%%%%%%%%%%%%
%\begin{table}[t!]
    %\vspace{-1mm}
	\footnotesize
	\captionof{table}{\small 
	Relative contribution of fusion network inputs
	%evaluated 
	on 3DPW, MPJPE ($\downarrow$).
	\vspace{-3.7mm}
	}
	\centering
	\setlength\tabcolsep{1.5pt}
	\resizebox{\textwidth}{!}{
	\begin{tabular}{l|c|c|c|c}
	%\toprule
	\hline
        Methods & $\hat{p}$ & $\tilde{q}$ & $\tilde{q}+\!\!\tilde{w}$ &  Fused  \\
        \hline%\hline
		\textit{Ours(H$\rightarrow$3PDW)} & 100 & 122 & {115} & \textbf{91}\\
		%\hline
		\textit{Ours(JU:H$\rightarrow$3DPW)} & 116 & 142 & 135 & \textbf{98}\\
		\hline
		%\bottomrule
	\end{tabular}}
	%\vspace{-2mm}
	\label{tab:fusion}
%\end{table} 
%%%%%%%%%%%%%%%%%%%%%%% Pre-adaptation table ends  %%%%%%%%%%%%%%%%%%%%%%%%%%%%%%%%%%%%%%%%%%%%%%%
\end{minipage}
%\hfill
% \begin{minipage}[t]{0.30\textwidth}
% %%%%%%%%%%%%%%%% Pre-adaptation table %%%%%%%%%%%%%%%%%%
% %\begin{table}[t!]
%     %\vspace{-1mm}
% 	\footnotesize
% 	\captionof{table}{\small
% 	2D pose advantage PCKh ($\uparrow$) 
% 	\vspace{-3mm}
% 	}
% 	\centering
% 	\setlength\tabcolsep{2.0pt}
% 	\resizebox{\textwidth}{!}{
% 	\begin{tabular}{l|c|c}
% 	%\toprule
% 	%\vspace{-2mm}
% 	\hline
%  		Methods & 3DPW & H3.6M  \\
% 		\hline
% 	    Pre-lift: \textit{PoseNet3D}, SH [\green{55}] & \textbf{81.2} & \textbf{85.7} \\
% 		Pre-adapt: \textit{Ours(SH$\rightarrow$ML)} & 68.8 & - \\
% 		%\hline
% 		Pre-adapt: \textit{Ours(S$\rightarrow$H)} & - & 57.6\\
% 		\hline
% 		%\bottomrule
% 	\end{tabular}}
% 	%\vspace{-2mm}
% 	\label{tab:preadapt}
% %\end{table} 
% %%%%%%%%%%%%%%%%%%%%%%% Pre-adaptation table ends  %%%%%%%%%%%%%%%%%%%%%%%%%%%%%%%%%%%%%%%%%%%%%%%
% \end{minipage}
\hfill
\begin{minipage}[t]{0.33\textwidth}
%%%%%%%%%%%%%%%%%%%%%%%%%% Hyperparam sensitivity figure %%%%%%%%%%%%%%%%%%%%%%%%%%%%
% \begin{figure}[!t]%[!tbp][h!]

 \begin{center}
 	%\vspace{-1mm}
 	\captionof{figure}{\small Hyperparameter sensitivity.
 	}
 	\vspace{-4mm}
 	\includegraphics[width=\linewidth]{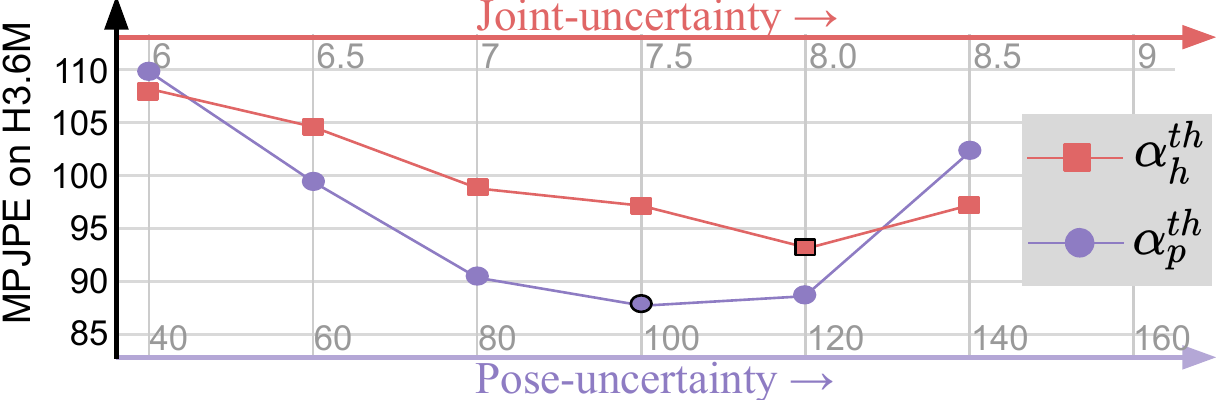}
 	%\vspace{-0.1mm}
     
     \label{fig:hyperparam_plot}  
 \end{center}
% \end{figure}
%%%%%%%%%%%%%%%%%%%%%%%%%% Hyperparam sensitivity figure ends %%%%%%%%%%%%%%%%%%%%%%%%%%%%
\end{minipage}

\end{table*}

%%%%%%%%%%%%%%%%%%%%%%%%%%%%%%%%%%%%%%%%%%%%%%%%%
\begin{figure*}[t] %[!b]%[!tbp][h!]
\begin{center}
    %\vspace{-1mm}
	\includegraphics[width=1.0\linewidth]{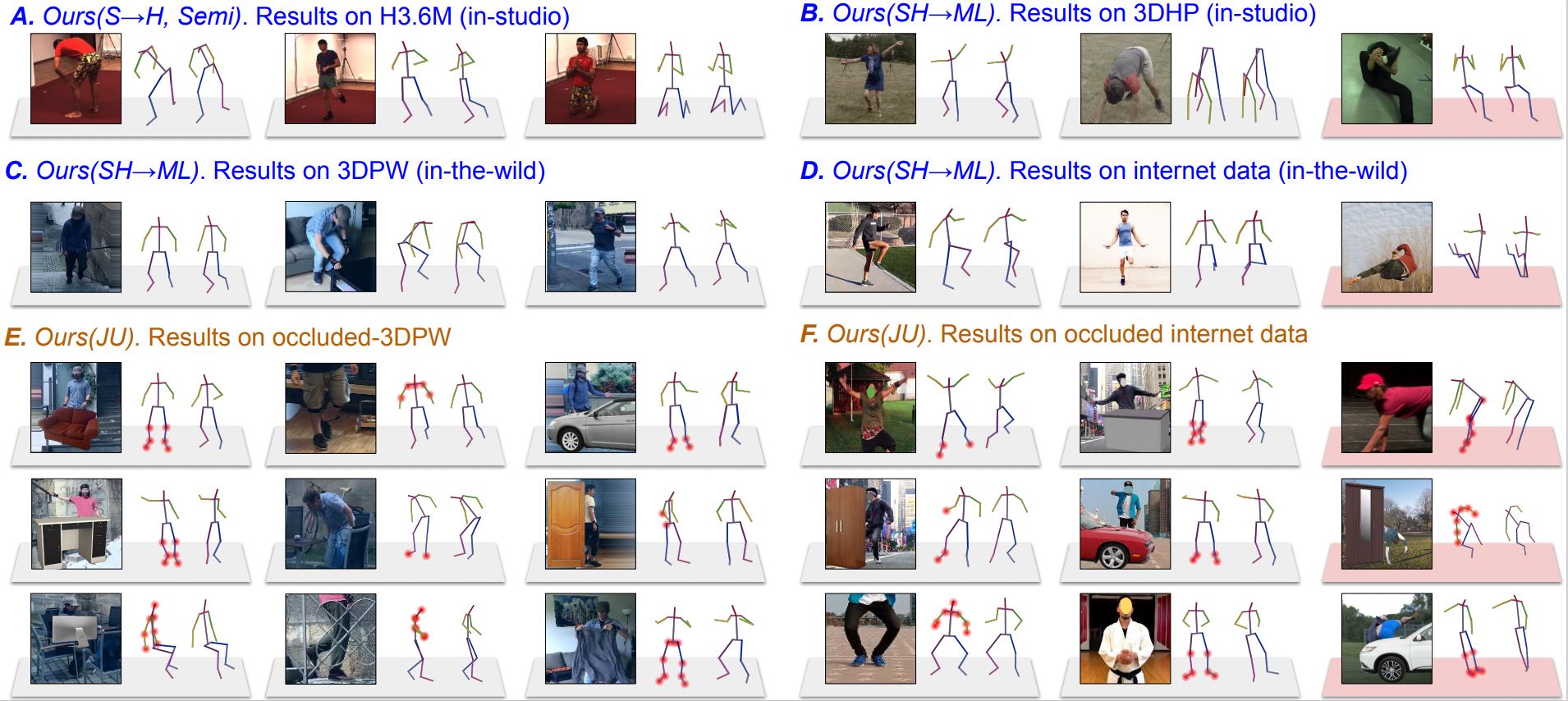}
	\vspace{-7mm}
	\caption{Qualitative analysis. 3D poses shown correspond to the original camera view and another azimuthal view %with azimuthal angle changed by
	at +30$^{\circ}$ or -30$^{\circ}$ depending on best viewing angle. %%% For full-body images the results are obtained via our best model for the corresponding datasets. %We use \textit{Ours(SH$\rightarrow$ML)} to obtain pose prediction for the in-the-wild internet data. 
	For results in panel \textbf{E} and \textbf{F} %depicts results from \textit{Ours(JU)} where 
	the joints with uncertainty greater than a prefix threshold are highlighted with red-blobs. The model fails on rare poses, complex inter-limb occlusion and heavy background clutter as highlighted by red bases.
	}
 	\label{fig:fig3}    
    %\vspace{-1mm}
\end{center}
\end{figure*}
%%%%%%%%%%%%%%%%% Fig Arch ends %%%%%%%%%%%%%%%%%

%%%%%%%%%%%%%%%%% Fig Arch %%%%%%%%%%%%%%%%%
\begin{figure*}[!h] %[!tbp]%[!b]%[!tbp][h!]
\begin{center}
    \vspace{-1mm}
	\includegraphics[width=1.0\linewidth]{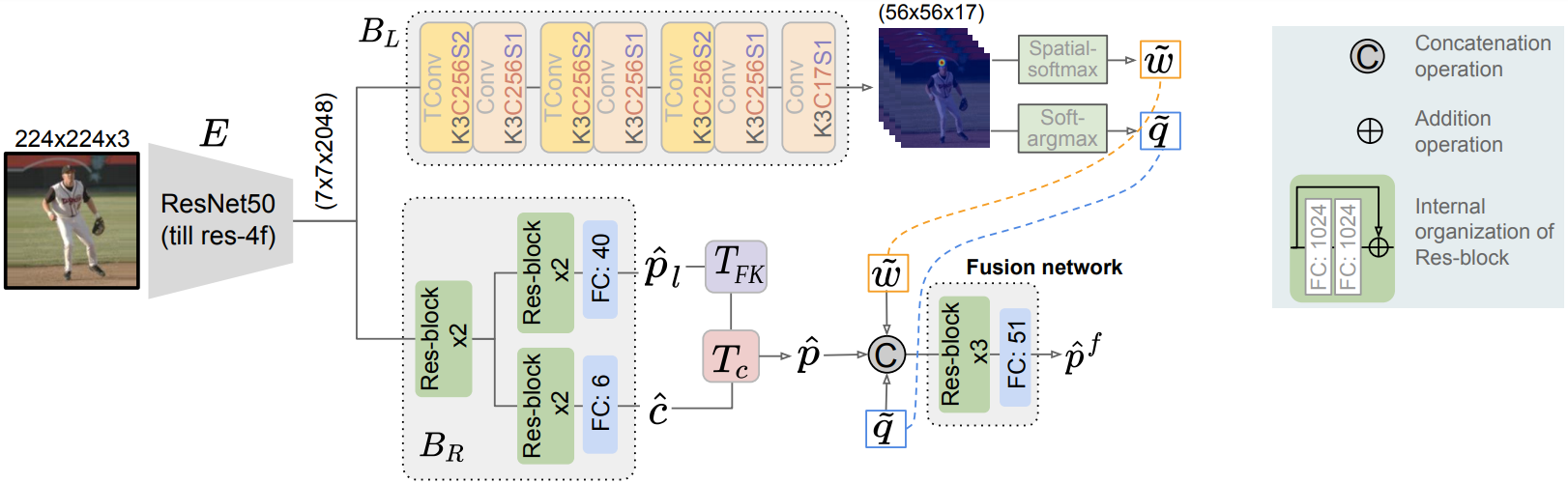}
	\vspace{-5mm}
	\caption{Detailed architecture of the proposed \textit{MRP-Net}. On the right we show the legend. Here, \textit{K3C256S2} denotes specifications of the convolutional layer, \ie 3$\times$3 filter size, 256 filters applied with a stride 2. Here, \textit{TConv} denotes transposed convolution operation. \textit{FC} denotes fully-connected layer. x2 and x3 depict number of residual blocks that are stacked to form the corresponding branch.
	}
 	\label{fig:arch}    
    \vspace{-2mm}
\end{center}
\end{figure*}
%%%%%%%%%%%%%%%%% Fig Arch ends %%%%%%%%%%%%%%%%%

%%%%%%%%%%%%%%%%% Fig Arch %%%%%%%%%%%%%%%%%
\begin{figure*}[!h]%[!b]%[!tbp][h!]
\begin{center}
    \vspace{-5mm}
	\includegraphics[width=1.0\linewidth]{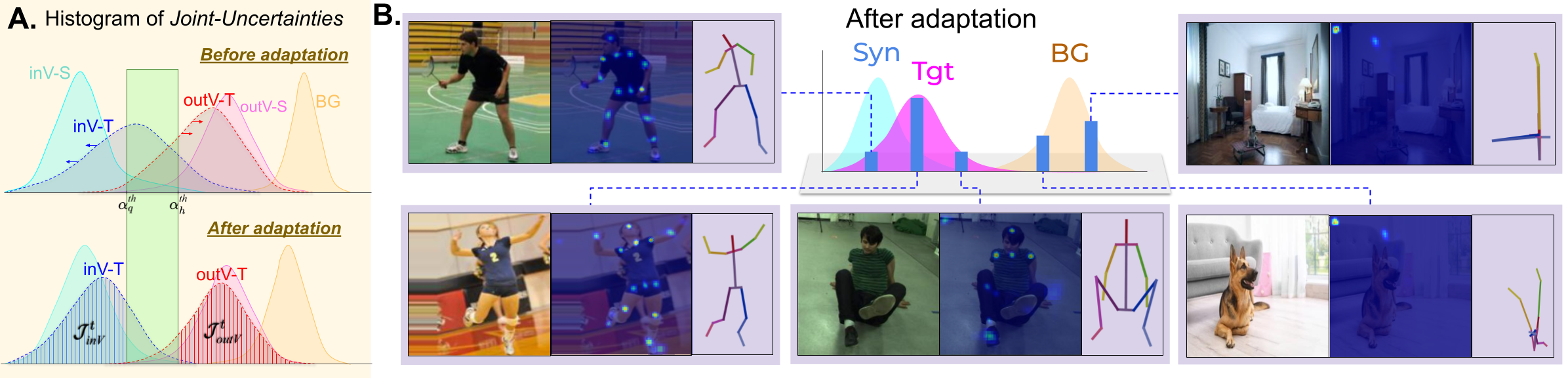}
	\vspace{-8mm}
	\caption{\textbf{A.} Shows histogram of the predicted \textit{joint-uncertainties} for the true \textit{in-view} and \textit{out-view} joints separately for source (\ie \texttt{inV-S} and \texttt{outV-S}) and target (\ie \texttt{inV-T} and \texttt{outV-T}). \texttt{BG} denotes the histogram of all \textit{out-view} joints for backgrounds. The \underline{shaded regions} in the bottom panel depicts $\mathcal{J}_\textit{inV}^t$ and $\mathcal{J}_\textit{outV}^t$ which are segregated using the preset thresholds $\alpha_q^\textit{th}$ and $\alpha_h^\textit{th}$ respectively (edges of the \underline{green-box}). Our adaptation algorithm succeeds to separate \textit{inV-T} and \textit{outV-T} over the course of adaptation training. 
	\textbf{B.} Shows a similar analysis for \textit{pose-uncertainties}. We show 5 different examples sampled from different regions of the histogram-bins. %%
	\textit{\textbf{Results on right-panel:}} Notice that to maximize \textit{pose-uncertainty} for backgrounds (OOD samples), \textit{MRPN} estimates the 2D landmarks and 3D pose points separated towards opposite diagonal corners. Here, the 2D landmarks are collapsed to the top-left corner whereas the root joint (pelvis) of the model-based 3D predictions are seemed to have collapsed towards the bottom-right corner. %%%
	\textit{\textbf{Result on bottom-panel:}} For uncertain target instances, we see two peaks in the joint heatmap PDFs; one at the top-left corner (OOD-related) and the other near the actual joint location. During adaptation, the OOD-related peak suppress while the joint-related peak rises to simultaneously reduce the uncertainty while converging towards the true pose outcome. %%%
	\textit{\textbf{Results on the left panel:}} \textit{Joint-level} uncertainty is indicated by the entropy of heatmap PDF.
	}
 	\label{fig:fig2}    
    \vspace{-3mm}
\end{center}
\end{figure*}
%%%%%%%%%%%%%%%%% Fig Arch ends %%%%%%%%%%%%%%%%%

%%%%%%%%%%%%%%%%% Fig Arch %%%%%%%%%%%%%%%%%
\begin{figure}[t] %[!tbp]%[!b]%[!tbp][h!]
\begin{center}
    %\vspace{-1mm}
	\includegraphics[width=1.0\linewidth]{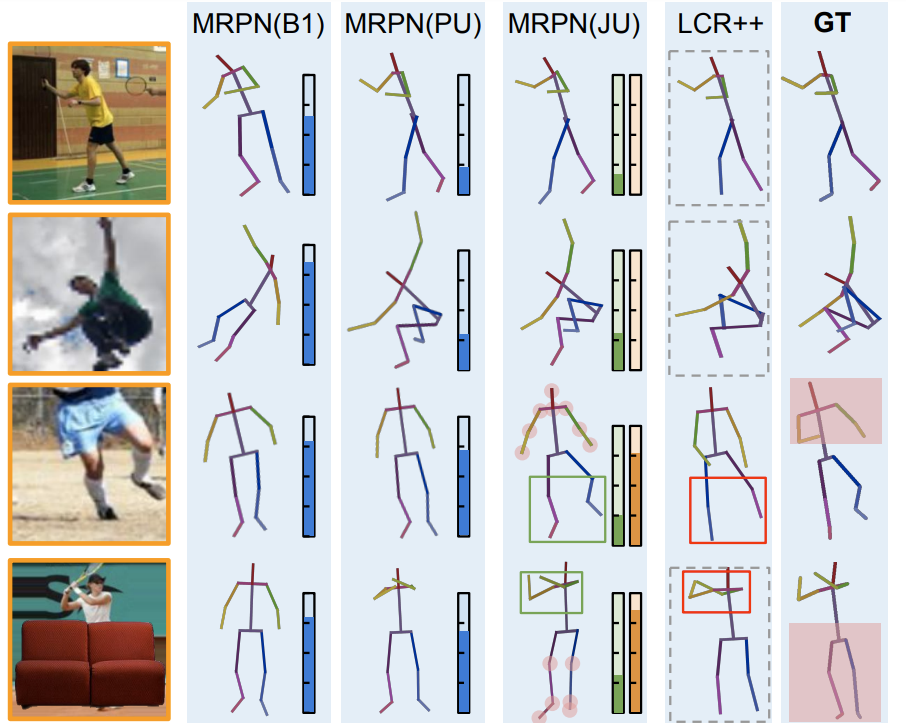}
	\vspace{-6mm}
	\caption{
	Every pose prediction of \textit{MRPN} is associated with a measure of uncertainty barometer. The barometer height indicates high uncertainty. The \underline{blue, green and orange barometers} indicate the average prediction uncertainty for the \texttt{full-pose}, \texttt{true-in-view} joints and \texttt{true-out-view} joints respectively. %
	%
	%%% \textit{MRPN(PU)} is not tuned to work on occluded/truncated images thus yields a higher uncertainty for the last two rows. Whereas, the uncertainty predictions of \textit{MRPN(JU)} for the green and orange barometer yield the expected behaviour. %
	%
	The \underline{dotted gray rectangles} highlight the failure cases of LCR++ in predicting the correct 3D inter-limb depth though the 2D landmarks align with the GT. 
	In the \underline{last 2 rows}, the filled red-box under GT column segregates the true \textit{out-view} joints. The \textit{in-view} joint predictions of \textit{MRPN(JU)}  (unfilled green rectangles) performs better against the same for LCR++ (unfilled red rectangles) when compared against the same under GT. 
	\vspace{-4mm}
	}
 	\label{fig:fig4}    
    \vspace{-4mm}
\end{center}
\end{figure}
%%%%%%%%%%%%%%%%% Fig Arch ends %%%%%%%%%%%%%%%%%

\section{Network architecture}
\label{sec:5}
The architecture consists of an ImageNet initialized ResNet-50 (till \textit{Res-4F}) which bifurcates into two branches, ${B}_L$ and ${B}_R$ as shown in Fig.~\ref{fig:arch}. ${B}_L$ is a convolutional decoder consisting of an alternate series of transposed convolution and general convolution which progressively increases the spatial resolution from 7$\times$7 to 56$\times$56. The final output of ${B}_L$ is 17 heatmap PDFs, $\tilde{h}$ obtained via spatial softmax. These are then used to extract the corresponding 2D joint coordinates, $\tilde{q}$ and joint confidence, $\tilde{w}$. ${B}_R$ consists of a common branch with fully-connected residual blocks~\cite{wandt2019repnet} which further divides into camera, $\hat{c}$ and pose prediction $\hat{p}_l$ sub-branches, each consisting of 2 residual blocks. The outputs, $\tilde{w}$, $\tilde{q}$, and $\hat{p}$ are concatenated and passed to the fusion network which is composed of a series of 3 residual blocks to regress the final 3D pose, $\hat{p}^f$. Fig.~\ref{fig:arch} shows the detailed architecture. Further, ablation performance with fusion network is shown in Table~\ref{tab:ablation_supp} (MPJPE of \#5-7, Table \textcolor{red}{4}). We see that a better adaptation further enhances the gain from fusion network.

% \section{Transforming local vectors to canonical pose}\label{sec:4}

% The pose branch from $B_R$ outputs 13 three dimensional unit vectors, one for each joint except the pelvis, hips and neck joints. These 13 joints are defined at their respective parent relative local coordinate system (\ie parent joint as the origin with axis directions obtained via Gram-Schmidt orthogonalization of the parent-limb vector and the \textit{face-vector})~\cite{akhter2015pose}. The pose branch also outputs the angle between the line joining the neck to the pelvis and the hip-line (line segment connecting the two hip joints).

% Here, $p^{3D}$ for \textit{pelvis} is set at origin, \ie $(0,0,0)$. Let $\textit{Pa}(j)$ represent parent of joint $j$ and $\alpha^{(\textit{Pa}(j)\rightarrow j)}$ be the fixed bone length of the line segment joining $\textit{Pa}(j)$ to $j$. The final recursive forward kinematic formulation is then given by,

% \vspace{-3mm}
% $$
% p^{3D}(j) = p^{3D}(\textit{Pa}(j)) + \alpha^{(\textit{Pa}(j)\rightarrow j)}g^{(j)}
% $$

% Here, $p^{3D}(j)$ is the position vector of joint $j$ in the 3D canonical coordinate system. $g^{(j)}$ is the unit vector normalization of $\tilde{g}^{(j)}$ which is the resultant direction in the canonical coordinate system, computed as:
% $$
% \tilde{g}^{(j)} = v_x^{3D}(j)a^{(j)} + v_y^{3D}(j)b + v_z^{3D}(j)n^{(j)}
% $$
% where the raw neural values for each joint $j$ are represented as $(v_x^{3D}(j), v_y^{3D}(j), v_z^{3D}(j))$.
% Finally, $p^{3D}: \{p^{3D}(j)\}_{j=1}^J$.

\section{Qualitative analysis}
\label{sec:7}
We perform a thorough qualitative study to interpret the behaviour of our network for a wide variety of in-distribution and out-of-distribution samples (see Fig.~\ref{fig:fig3}). %In Fig.~\ref{fig:fig2}, we analyze how samples from different domains are distributed along the uncertainty metrics, \ie the \textit{pose-uncertainty} and \textit{joint-uncertainty}. Fig.~\ref{fig:fig3} and Fig.~\ref{fig:fig4} analyze our pose prediction results across variations in pose complexity, occlusion/truncation scenarios and environmental conditions (\ie in-studio and in-the-wild). 

The analysis in Fig.~\ref{fig:fig2}{\color{red}A} shows that the proposed joint-level adaptation algorithm succeeds to separate \textit{inV-T} and \textit{outV-T} over the course of adaptation training, thereby aligning these with \textit{inV-S} and \textit{outV-S} respectively. 
In Fig.~\ref{fig:fig4}, \textit{MRPN(B1)} indicates the occlusion-aware network before the adaptation training. \textit{MRPN(PU)} and \textit{MRPN(JU)} indicate the final networks after the \textit{pose-level} and \textit{joint-level} adaptations. %respectively. 
Further we show the ground-truth (2D) and predictions on LCR++~\cite{rogez2019lcr}. \textit{MRPN(PU)} is not tuned to work on occluded/truncated images and thus yields a higher uncertainty for the last two rows. Whereas, the uncertainty predictions of \textit{MRPN(JU)} for the green and orange barometer yield the expected behaviour.

\begin{table}[!t]
    %\vspace{-1mm}
	%\footnotesize
	%\vspace{-3mm}
	\caption{Evaluation of \#5-7 from Table 4 with fusion network.
	\vspace{-3mm}}
	\centering
	\setlength\tabcolsep{1.5pt}
	\resizebox{\linewidth}{!}{
	\begin{tabular}{l|l|c|c|c|c|c}
	\hline

		%\multicolumn{6}{c}{\texttt{Pose-level adaptation on {Human3.6M}}}\\ \hline
 		No. & Method & \makecell{$\mathcal{L}_\textit{Sup}^{(s)} - \mathcal{U}^{(b)}$}& $\mathcal{U}^t$ & $\mathcal{L}_\textit{pSup}^{(t)}$ & w/o fuse & w/ fuse \\
		\hline \hline
		5. & {\scriptsize \textit{B2(S$\rightarrow$H)+DANN}}
		& only $\mathcal{L}_\textit{Sup}^{(s)}$ & 
		\multicolumn{2}{c|}{Standard DA} & 116.8 & {114.5} \textcolor{dark_green}{(2.3 $\downarrow$)} \\
		\cline{3-5}
		6. & \textit{B2(S$\rightarrow$H)} & \cmark & - & - & 122.4 & {122.1} \textcolor{dark_green}{(0.3 $\downarrow$)}\\

		7. & \textit{B2(S$\rightarrow$H)} & \cmark & \cmark & - & 113.4  & {110.7} \textcolor{dark_green}{(2.7 $\downarrow$)}\\
		\bottomrule
	\end{tabular}}
	\vspace{-4mm}
	\label{tab:ablation_supp}
\end{table}
%%%%%%%%%%%%%%%%%%%%%%%%%%%%%%%%

%\pagebreak
\vspace{4mm}
{\small
\bibliographystyle{ieee_fullname}
\bibliography{egbib}

\begin{thebibliography}{100}\itemsep=-1pt

\bibitem{1542030}
Ankur Agarwal and Bill Triggs.
\newblock Recovering {3D} human pose from monocular images.
\newblock {\em IEEE transactions on pattern analysis and machine intelligence},
  28(1):44--58, 2005.

\bibitem{andriluka14cvpr}
Mykhaylo Andriluka, Leonid Pishchulin, Peter Gehler, and Bernt Schiele.
\newblock {2D} human pose estimation: New benchmark and state of the art
  analysis.
\newblock In {\em CVPR}, 2014.

\bibitem{6909612}
V. {Belagiannis}, S. {Amin}, M. {Andriluka}, B. {Schiele}, N. {Navab}, and S.
  {Ilic}.
\newblock {3D} pictorial structures for multiple human pose estimation.
\newblock In {\em CVPR}, 2014.

\bibitem{5206699}
L. {Bo} and C. {Sminchisescu}.
\newblock Structured output-associative regression.
\newblock In {\em CVPR}, 2009.

\bibitem{Bogo:ECCV:2016}
Federica Bogo, Angjoo Kanazawa, Christoph Lassner, Peter Gehler, Javier Romero,
  and Michael~J. Black.
\newblock Keep it {SMPL}: Automatic estimation of {3D} human pose and shape
  from a single image.
\newblock In {\em ECCV}, 2016.

\bibitem{5206523}
P. {Buehler}, A. {Zisserman}, and M. {Everingham}.
\newblock Learning sign language by watching tv (using weakly aligned
  subtitles).
\newblock In {\em CVPR}, 2009.

\bibitem{6619308}
M. {Burenius}, J. {Sullivan}, and S. {Carlsson}.
\newblock {3D} pictorial structures for multiple view articulated pose
  estimation.
\newblock In {\em CVPR}, 2013.

\bibitem{Burge16849}
Johannes Burge and Wilson~S Geisler.
\newblock Optimal defocus estimation in individual natural images.
\newblock {\em Proceedings of the National Academy of Sciences},
  108(40):16849--16854, 2011.

\bibitem{cao2019cross}
Jinkun Cao, Hongyang Tang, Hao-Shu Fang, Xiaoyong Shen, Cewu Lu, and Yu-Wing
  Tai.
\newblock Cross-domain adaptation for animal pose estimation.
\newblock In {\em ICCV}, 2019.

\bibitem{chen2019unsupervised}
Ching-Hang Chen, Ambrish Tyagi, Amit Agrawal, Dylan Drover, Rohith MV, Stefan
  Stojanov, and James~M Rehg.
\newblock Unsupervised {3D} pose estimation with geometric self-supervision.
\newblock In {\em CVPR}, 2019.

\bibitem{chen2019weakly}
Xipeng Chen, Kwan-Yee Lin, Wentao Liu, Chen Qian, and Liang Lin.
\newblock Weakly-supervised discovery of geometry-aware representation for {3D}
  human pose estimation.
\newblock In {\em CVPR}, 2019.

\bibitem{chen2020eccv}
Zerui Chen, Yan Huang, Hongyuan Yu, Bin Xue, Ke Han, Yiru Guo, and Liang Wang.
\newblock Towards part-aware monocular {3D} human pose estimation: An
  architecture search approach.
\newblock In {\em ECCV}, 2020.

\bibitem{cheng20203d}
Yu Cheng, Bo Yang, Bo Wang, and Robby~T Tan.
\newblock {3D} human pose estimation using spatio-temporal networks with
  explicit occlusion training.
\newblock In {\em AAAI}, 2020.

\bibitem{cheng2019occlusion}
Yu Cheng, Bo Yang, Bo Wang, Wending Yan, and Robby~T Tan.
\newblock Occlusion-aware networks for {3D} human pose estimation in video.
\newblock In {\em ICCV}, 2019.

\bibitem{Dabral_2018_ECCV}
Rishabh Dabral, Anurag Mundhada, Uday Kusupati, Safeer Afaque, Abhishek Sharma,
  and Arjun Jain.
\newblock Learning {3D} human pose from structure and motion.
\newblock In {\em ECCV}, 2018.

\bibitem{de2018deep}
Rodrigo de Bem, Anurag Arnab, Stuart Golodetz, Michael Sapienza, and Philip
  Torr.
\newblock Deep fully-connected part-based models for human pose estimation.
\newblock In {\em ACML}, 2018.

\bibitem{NIPS2019_9454}
Carl Doersch and Andrew Zisserman.
\newblock Sim2real transfer learning for {3D} human pose estimation: motion to
  the rescue.
\newblock In {\em NeurIPS}, 2019.

\bibitem{gal2016dropout}
Yarin Gal and Zoubin Ghahramani.
\newblock Dropout as a bayesian approximation: Representing model uncertainty
  in deep learning.
\newblock In {\em ICML}, 2016.

\bibitem{ganin2015unsupervised}
Yaroslav Ganin and Victor Lempitsky.
\newblock Unsupervised domain adaptation by backpropagation.
\newblock In {\em ICML}, 2015.

\bibitem{ganin2016domain}
Yaroslav Ganin, Evgeniya Ustinova, Hana Ajakan, Pascal Germain, Hugo
  Larochelle, Fran{\c{c}}ois Laviolette, Mario Marchand, and Victor Lempitsky.
\newblock Domain-adversarial training of neural networks.
\newblock {\em The journal of machine learning research}, 17(1):2096--2030,
  2016.

\bibitem{10.1167/11.12.14}
Wilson~S Geisler and Jeffrey~S Perry.
\newblock Statistics for optimal point prediction in natural images.
\newblock {\em Journal of Vision}, 11(12):14--14, 2011.

\bibitem{gower1975generalized}
John~C Gower.
\newblock Generalized procrustes analysis.
\newblock {\em Psychometrika}, 40(1):33--51, 1975.

\bibitem{Guler_2019_CVPR}
Riza~Alp Guler and Iasonas Kokkinos.
\newblock Holopose: Holistic {3D} human reconstruction in-the-wild.
\newblock In {\em CVPR}, 2019.

\bibitem{hagbi2010shape}
Nate Hagbi, Oriel Bergig, Jihad El-Sana, and Mark Billinghurst.
\newblock Shape recognition and pose estimation for mobile augmented reality.
\newblock {\em IEEE transactions on visualization and computer graphics},
  17(10):1369--1379, 2010.

\bibitem{he2016deep}
Kaiming He, Xiangyu Zhang, Shaoqing Ren, and Jian Sun.
\newblock Deep residual learning for image recognition.
\newblock In {\em CVPR}, 2016.

\bibitem{hendrycks2016baseline}
Dan Hendrycks and Kevin Gimpel.
\newblock A baseline for detecting misclassified and out-of-distribution
  examples in neural networks.
\newblock In {\em ICLR}, 2017.

\bibitem{10.1007/978-3-030-01249-6_5}
Mir Rayat~Imtiaz Hossain and James~J. Little.
\newblock Exploiting temporal information for {3D} human pose estimation.
\newblock In {\em ECCV}, 2018.

\bibitem{ionescu2013human3}
Catalin Ionescu, Dragos Papava, Vlad Olaru, and Cristian Sminchisescu.
\newblock Human3.6m: Large scale datasets and predictive methods for {3D} human
  sensing in natural environments.
\newblock {\em IEEE transactions on pattern analysis and machine intelligence},
  36(7):1325--1339, 2013.

\bibitem{Iqbal_2020_CVPR}
Umar Iqbal, Pavlo Molchanov, and Jan Kautz.
\newblock Weakly-supervised {3D} human pose learning via multi-view images in
  the wild.
\newblock In {\em CVPR}, 2020.

\bibitem{kanazawa2018end}
Angjoo Kanazawa, Michael~J Black, David~W Jacobs, and Jitendra Malik.
\newblock End-to-end recovery of human shape and pose.
\newblock In {\em CVPR}, 2018.

\bibitem{kingma2014adam}
Diederik~P Kingma and Jimmy Ba.
\newblock Adam: A method for stochastic optimization.
\newblock In {\em ICLR}, 2015.

\bibitem{kocabas2019self}
Muhammed Kocabas, Salih Karagoz, and Emre Akbas.
\newblock Self-supervised learning of {3D} human pose using multi-view
  geometry.
\newblock In {\em CVPR}, 2019.

\bibitem{kolotouros2019learning}
Nikos Kolotouros, Georgios Pavlakos, Michael~J Black, and Kostas Daniilidis.
\newblock Learning to reconstruct {3D} human pose and shape via model-fitting
  in the loop.
\newblock In {\em ICCV}, 2019.

\bibitem{kong2019deep}
Chen Kong and Simon Lucey.
\newblock Deep non-rigid structure from motion.
\newblock In {\em ICCV}, 2019.

\bibitem{kuleshov2018accurate}
Volodymyr Kuleshov, Nathan Fenner, and Stefano Ermon.
\newblock Accurate uncertainties for deep learning using calibrated regression.
\newblock In {\em ICML}, 2018.

\bibitem{kundu2020cross}
Jogendra~Nath Kundu, Himanshu Buckchash, Priyanka Mandikal, Rahul MV, Anirudh
  Jamkhandi, and R~Venkatesh Babu.
\newblock Cross-conditioned recurrent networks for long-term synthesis of
  inter-person human motion interactions.
\newblock In {\em WACV}, 2020.

\bibitem{kundu2019_um_adapt}
Jogendra~Nath Kundu, Nishank Lakkakula, and R~Venkatesh Babu.
\newblock Um-adapt: Unsupervised multi-task adaptation using adversarial
  cross-task distillation.
\newblock In {\em ICCV}, 2019.

\bibitem{kundu2018object}
Jogendra~Nath Kundu, Rahul MV, Aditya Ganeshan, and R.~Venkatesh Babu.
\newblock Object pose estimation from monocular image using multi-view keypoint
  correspondence.
\newblock In {\em ECCV Workshops}, 2018.

\bibitem{Kundu_2020_WACV}
Jogendra~Nath Kundu, Rahul MV, Jay Patravali, and R~Venkatesh Babu.
\newblock Unsupervised cross-dataset adaptation via probabilistic amodal {3D}
  human pose completion.
\newblock In {\em WACV}, 2020.

\bibitem{kundu_appearance}
Jogendra~Nath Kundu, Mugalodi Rakesh, Varun Jampani, Rahul MV, and R.~Venkatesh
  Babu.
\newblock Appearance consensus driven self-supervised human mesh recovery.
\newblock In {\em ECCV}, 2020.

\bibitem{kundu2020unsup}
Jogendra~Nath Kundu, Ambareesh Revanur, Govind~V Waghmare, Rahul MV, and
  R~Venkatesh Babu.
\newblock Unsupervised cross-modal alignment for multi-person {3D} pose
  estimation.
\newblock In {\em ECCV}, 2020.

\bibitem{kundu2021non}
Jogendra~Nath Kundu, Siddharth Seth, Anirudh Jamkhandi, Pradyumna YM, Varun
  Jampani, Anirban Chakraborty, and R~Venkatesh Babu.
\newblock Non-local latent relation distillation for self-adaptive {3D} human
  pose estimation.
\newblock In {\em NeurIPS}, 2021.

\bibitem{kundu2020self}
Jogendra~Nath Kundu, Siddharth Seth, Varun Jampani, Mugalodi Rakesh,
  R~Venkatesh Babu, and Anirban Chakraborty.
\newblock Self-supervised {3D} human pose estimation via part guided novel
  image synthesis.
\newblock In {\em CVPR}, 2020.

\bibitem{kundu2020ksp}
Jogendra~Nath Kundu, Siddharth Seth, Rahul MV, Rakesh Mugalodi, R~Venkatesh
  Babu, and Anirban Chakraborty.
\newblock Kinematic-structure-preserved representation for unsupervised {3D}
  human pose estimation.
\newblock In {\em AAAI}, 2020.

\bibitem{lakshminarayanan2016simple}
Balaji Lakshminarayanan, Alexander Pritzel, and Charles Blundell.
\newblock Simple and scalable predictive uncertainty estimation using deep
  ensembles.
\newblock In {\em NeurIPS}, 2017.

\bibitem{lee2013pseudo}
Dong-Hyun Lee et~al.
\newblock Pseudo-label: The simple and efficient semi-supervised learning
  method for deep neural networks.
\newblock In {\em ICML Workshops}, 2013.

\bibitem{lee2018simple}
Kimin Lee, Kibok Lee, Honglak Lee, and Jinwoo Shin.
\newblock A simple unified framework for detecting out-of-distribution samples
  and adversarial attacks.
\newblock In {\em ICLR}, 2018.

\bibitem{li2017deeper}
Da Li, Yongxin Yang, Yi-Zhe Song, and Timothy~M Hospedales.
\newblock Deeper, broader and artier domain generalization.
\newblock In {\em ICCV}, 2017.

\bibitem{Li_2020_CVPR}
Shichao Li, Lei Ke, Kevin Pratama, Yu-Wing Tai, Chi-Keung Tang, and Kwang-Ting
  Cheng.
\newblock Cascaded deep monocular {3D} human pose estimation with evolutionary
  training data.
\newblock In {\em CVPR}, 2020.

\bibitem{9010633}
Z. {Li}, X. {Wang}, F. {Wang}, and P. {Jiang}.
\newblock On boosting single-frame {3D} human pose estimation via monocular
  videos.
\newblock In {\em ICCV}, 2019.

\bibitem{liang2017enhancing}
Shiyu Liang, Yixuan Li, and Rayadurgam Srikant.
\newblock Enhancing the reliability of out-of-distribution image detection in
  neural networks.
\newblock In {\em ICLR}, 2018.

\bibitem{long2015learning}
Mingsheng Long, Yue Cao, Jianmin Wang, and Michael Jordan.
\newblock Learning transferable features with deep adaptation networks.
\newblock In {\em ICML}, 2015.

\bibitem{martinez2017simple}
Julieta Martinez, Rayat Hossain, Javier Romero, and James~J Little.
\newblock A simple yet effective baseline for {3D} human pose estimation.
\newblock In {\em ICCV}, 2017.

\bibitem{mehta2017monocular}
Dushyant Mehta, Helge Rhodin, Dan Casas, Pascal Fua, Oleksandr Sotnychenko,
  Weipeng Xu, and Christian Theobalt.
\newblock Monocular {3D} human pose estimation in the wild using improved cnn
  supervision.
\newblock In {\em {3D}V}, 2017.

\bibitem{mehta2018single}
Dushyant Mehta, Oleksandr Sotnychenko, Franziska Mueller, Weipeng Xu, Srinath
  Sridhar, Gerard Pons-Moll, and Christian Theobalt.
\newblock Single-shot multi-person {3D} pose estimation from monocular rgb.
\newblock In {\em 3DV}, 2018.

\bibitem{Mehta2017vnect}
Dushyant Mehta, Srinath Sridhar, Oleksandr Sotnychenko, Helge Rhodin, Mohammad
  Shafiei, Hans-Peter Seidel, Weipeng Xu, Dan Casas, and Christian Theobalt.
\newblock {VNect}: Real-time {3D} human pose estimation with a single rgb
  camera.
\newblock {\em ACM Transactions on Graphics (TOG)}, 36(4):1--14, 2017.

\bibitem{Mitra_2020_CVPR}
Rahul Mitra, Nitesh~B. Gundavarapu, Abhishek Sharma, and Arjun Jain.
\newblock Multiview-consistent semi-supervised learning for {3D} human pose
  estimation.
\newblock In {\em CVPR}, 2020.

\bibitem{moon2020i2l}
Gyeongsik Moon and Kyoung~Mu Lee.
\newblock {I2L-MeshNet}: Image-to-lixel prediction network for accurate {3D}
  human pose and mesh estimation from a single rgb image.
\newblock In {\em ECCV}, 2020.

\bibitem{moreno20173d}
Francesc Moreno-Noguer.
\newblock {3D} human pose estimation from a single image via distance matrix
  regression.
\newblock In {\em CVPR}, 2017.

\bibitem{9157335}
Jiteng Mu, Weichao Qiu, Gregory~D. Hager, and Alan~L. Yuille.
\newblock Learning from synthetic animals.
\newblock In {\em CVPR}, 2020.

\bibitem{newell2016stacked}
Alejandro Newell, Kaiyu Yang, and Jia Deng.
\newblock Stacked hourglass networks for human pose estimation.
\newblock In {\em ECCV}, 2016.

\bibitem{8237635}
B.~X. {Nie}, P. {Wei}, and S. {Zhu}.
\newblock Monocular {3D} human pose estimation by predicting depth on joints.
\newblock In {\em ICCV}, 2017.

\bibitem{novotny2019c3dpo}
David Novotny, Nikhila Ravi, Benjamin Graham, Natalia Neverova, and Andrea
  Vedaldi.
\newblock {C3DPO}: Canonical {3D} pose networks for non-rigid structure from
  motion.
\newblock In {\em ICCV}, 2019.

\bibitem{SMPL-X:2019}
Georgios Pavlakos, Vasileios Choutas, Nima Ghorbani, Timo Bolkart, Ahmed A.~A.
  Osman, Dimitrios Tzionas, and Michael~J. Black.
\newblock Expressive body capture: {3D} hands, face, and body from a single
  image.
\newblock In {\em CVPR}, 2019.

\bibitem{pavlakos2018ordinal}
Georgios Pavlakos, Xiaowei Zhou, and Kostas Daniilidis.
\newblock Ordinal depth supervision for {3D} human pose estimation.
\newblock In {\em CVPR}, 2018.

\bibitem{8099622}
G. {Pavlakos}, X. {Zhou}, K.~G. {Derpanis}, and K. {Daniilidis}.
\newblock Coarse-to-fine volumetric prediction for single-image {3D} human
  pose.
\newblock In {\em CVPR}, 2017.

\bibitem{pavllo2018quaternet}
Dario Pavllo, David Grangier, and Michael Auli.
\newblock Quaternet: A quaternion-based recurrent model for human motion.
\newblock In {\em BMVC}, 2018.

\bibitem{radwan2013monocular}
Ibrahim Radwan, Abhinav Dhall, and Roland Goecke.
\newblock Monocular image {3D} human pose estimation under self-occlusion.
\newblock In {\em ICCV}, 2013.

\bibitem{rakesh2021aligning}
Mugalodi Rakesh, Jogendra~Nath Kundu, Varun Jampani, and R~Venkatesh Babu.
\newblock Aligning silhouette topology for self-adaptive {3D} human pose
  recovery.
\newblock {\em NeurIPS}, 2021.

\bibitem{rhodin2018unsupervised}
Helge Rhodin, Mathieu Salzmann, and Pascal Fua.
\newblock Unsupervised geometry-aware representation for {3D} human pose
  estimation.
\newblock In {\em ECCV}, 2018.

\bibitem{rhodin2018learning}
Helge Rhodin, J{\"o}rg Sp{\"o}rri, Isinsu Katircioglu, Victor Constantin,
  Fr{\'e}d{\'e}ric Meyer, Erich M{\"u}ller, Mathieu Salzmann, and Pascal Fua.
\newblock Learning monocular {3D} human pose estimation from multi-view images.
\newblock In {\em CVPR}, 2018.

\bibitem{Guler2018DensePose}
Iasonas~Kokkinos Riza Alp~Guler, Natalia~Neverova.
\newblock Densepose: Dense human pose estimation in the wild.
\newblock In {\em CVPR}, 2018.

\bibitem{rogez2017lcr}
Gregory Rogez, Philippe Weinzaepfel, and Cordelia Schmid.
\newblock {LCR-Net}: Localization-classification-regression for human pose.
\newblock In {\em CVPR}, 2017.

\bibitem{rogez2019lcr}
Gregory Rogez, Philippe Weinzaepfel, and Cordelia Schmid.
\newblock {LCR-Net++}: Multi-person {2D} and {3D} pose detection in natural
  images.
\newblock {\em IEEE transactions on pattern analysis and machine intelligence},
  42(5):1146--1161, 2019.

\bibitem{Rosales2001LearningBP}
R{\'o}mer Rosales and S. Sclaroff.
\newblock Learning body pose via specialized maps.
\newblock In {\em NeurIPS}, 2001.

\bibitem{saito2017asymmetric}
Kuniaki Saito, Yoshitaka Ushiku, and Tatsuya Harada.
\newblock Asymmetric tri-training for unsupervised domain adaptation.
\newblock In {\em ICML}, 2017.

\bibitem{syguan2021boa}
Guan Shanyan, Xu Jingwei, Wang Yunbo, Ni Bingbing, and Yang Xiaokang.
\newblock Bilevel online adaptation for out-of-domain human mesh
  reconstruction.
\newblock In {\em CVPR}, 2021.

\bibitem{sigal2010humaneva}
Leonid Sigal, Alexandru~O Balan, and Michael~J Black.
\newblock Humaneva: Synchronized video and motion capture dataset and baseline
  algorithm for evaluation of articulated human motion.
\newblock {\em International journal of computer vision}, 87(1):4--27, 2010.

\bibitem{sun2019deep}
Ke Sun, Bin Xiao, Dong Liu, and Jingdong Wang.
\newblock Deep high-resolution representation learning for human pose
  estimation.
\newblock In {\em CVPR}, 2019.

\bibitem{sun2018integral}
Xiao Sun, Bin Xiao, Fangyin Wei, Shuang Liang, and Yichen Wei.
\newblock Integral human pose regression.
\newblock In {\em ECCV}, 2018.

\bibitem{tripathi2020posenet3d}
Shashank Tripathi, Siddhant Ranade, Ambrish Tyagi, and Amit Agrawal.
\newblock {PoseNet3D}: Learning temporally consistent {3D} human pose via
  knowledge distillation.
\newblock In {\em 3DV}, 2020.

\bibitem{TrumbleBMVC2017}
Matt Trumble, Andrew Gilbert, Charles Malleson, Adrian Hilton, and John
  Collomosse.
\newblock Total capture: {3D} human pose estimation fusing video and inertial
  sensors.
\newblock In {\em BMVC}, 2017.

\bibitem{tung2017adversarial}
Hsiao-Yu~Fish Tung, Adam~W Harley, William Seto, and Katerina Fragkiadaki.
\newblock Adversarial inverse graphics networks: Learning {2D}-to-{3D} lifting
  and image-to-image translation from unpaired supervision.
\newblock In {\em ICCV}, 2017.

\bibitem{tzeng2017adversarial}
Eric Tzeng, Judy Hoffman, Kate Saenko, and Trevor Darrell.
\newblock Adversarial discriminative domain adaptation.
\newblock In {\em CVPR}, 2017.

\bibitem{Tzeng2014DeepDC}
Eric Tzeng, Judy Hoffman, Ning Zhang, Kate Saenko, and Trevor Darrell.
\newblock Deep domain confusion: Maximizing for domain invariance.
\newblock {\em ArXiv}, abs/1412.3474, 2014.

\bibitem{varol2017learning}
Gul Varol, Javier Romero, Xavier Martin, Naureen Mahmood, Michael~J Black, Ivan
  Laptev, and Cordelia Schmid.
\newblock Learning from synthetic humans.
\newblock In {\em CVPR}, 2017.

\bibitem{von2018recovering}
Timo Von~Marcard, Roberto Henschel, Michael~J Black, Bodo Rosenhahn, and Gerard
  Pons-Moll.
\newblock Recovering accurate {3D} human pose in the wild using imus and a
  moving camera.
\newblock In {\em ECCV}, 2018.

\bibitem{wandt2019repnet}
Bastian Wandt and Bodo Rosenhahn.
\newblock Repnet: Weakly supervised training of an adversarial reprojection
  network for {3D} human pose estimation.
\newblock In {\em CVPR}, 2019.

\bibitem{wulfmeier2018incremental}
Markus Wulfmeier, Alex Bewley, and Ingmar Posner.
\newblock Incremental adversarial domain adaptation for continually changing
  environments.
\newblock In {\em ICRA}, 2018.

\bibitem{Xu_2020_CVPR}
Jingwei Xu, Zhenbo Yu, Bingbing Ni, Jiancheng Yang, Xiaokang Yang, and Wenjun
  Zhang.
\newblock Deep kinematics analysis for monocular {3D} human pose estimation.
\newblock In {\em CVPR}, 2020.

\bibitem{8578649}
W. {Yang}, W. {Ouyang}, X. {Wang}, J. {Ren}, H. {Li}, and X. {Wang}.
\newblock {3D} human pose estimation in the wild by adversarial learning.
\newblock In {\em CVPR}, 2018.

\bibitem{yu15lsun}
Fisher Yu, Yinda Zhang, Shuran Song, Ari Seff, and Jianxiong Xiao.
\newblock {LSUN}: Construction of a large-scale image dataset using deep
  learning with humans in the loop.
\newblock {\em arXiv preprint arXiv:1506.03365}, 2015.

\bibitem{6710175}
Amir~Roshan Zamir and Mubarak Shah.
\newblock Image geo-localization based on multiple nearest neighbor feature
  matching using generalized graphs.
\newblock {\em IEEE transactions on pattern analysis and machine intelligence},
  36(8):1546--1558, 2014.

\bibitem{zanfir2018monocular}
Andrei Zanfir, Elisabeta Marinoiu, and Cristian Sminchisescu.
\newblock Monocular {3D} pose and shape estimation of multiple people in
  natural scenes-the importance of multiple scene constraints.
\newblock In {\em CVPR}, 2018.

\bibitem{zanfir2018deep}
Andrei Zanfir, Elisabeta Marinoiu, Mihai Zanfir, Alin-Ionut Popa, and Cristian
  Sminchisescu.
\newblock Deep network for the integrated {3D} sensing of multiple people in
  natural images.
\newblock {\em NeurIPS}, 2018.

\bibitem{inference_stage2020}
Jianfeng Zhang, Xuecheng Nie, and Jiashi Feng.
\newblock Inference stage optimization for cross-scenario {3D} human pose
  estimation.
\newblock In {\em NeurIPS}, 2020.

\bibitem{mm_domain19}
Xiheng Zhang, Yongkang Wong, Mohan~S. Kankanhalli, and Weidong Geng.
\newblock Unsupervised domain adaptation for {3D} human pose estimation.
\newblock In {\em ACMMM}, 2019.

\bibitem{Zhang_2021_ICCV}
Xiheng Zhang, Yongkang Wong, Xiaofei Wu, Juwei Lu, Mohan Kankanhalli, Xiangdong
  Li, and Weidong Geng.
\newblock Learning causal representation for training cross-domain pose
  estimator via generative interventions.
\newblock In {\em ICCV}, 2021.

\bibitem{zhaoCVPR19semantic}
Long Zhao, Xi Peng, Yu Tian, Mubbasir Kapadia, and Dimitris~N. Metaxas.
\newblock Semantic graph convolutional networks for {3D} human pose regression.
\newblock In {\em CVPR}, 2019.

\bibitem{Zhou_2017_ICCV}
Xingyi Zhou, Qixing Huang, Xiao Sun, Xiangyang Xue, and Yichen Wei.
\newblock Towards {3D} human pose estimation in the wild: A weakly-supervised
  approach.
\newblock In {\em ICCV}, 2017.

\bibitem{zhou2016deep}
Xingyi Zhou, Xiao Sun, Wei Zhang, Shuang Liang, and Yichen Wei.
\newblock Deep kinematic pose regression.
\newblock In {\em ECCV}, 2016.

\bibitem{7780906}
X. {Zhou}, M. {Zhu}, S. {Leonardos}, K.~G. {Derpanis}, and K. {Daniilidis}.
\newblock Sparseness meets deepness: {3D} human pose estimation from monocular
  video.
\newblock In {\em CVPR}, 2016.

\bibitem{zou2018unsupervised}
Yang Zou, Zhiding Yu, BVK Kumar, and Jinsong Wang.
\newblock Unsupervised domain adaptation for semantic segmentation via
  class-balanced self-training.
\newblock In {\em ECCV}, 2018.

\end{thebibliography}
}

\end{document}